\documentclass[10pt,twocolumn,letterpaper]{article}

\usepackage{cvpr}              

\usepackage{graphicx}
\usepackage{amsmath}
\usepackage{amssymb}
\usepackage{booktabs}

\usepackage[pagebackref,breaklinks,colorlinks]{hyperref}

\usepackage[capitalize]{cleveref}
\crefname{section}{Sec.}{Secs.}
\Crefname{section}{Section}{Sections}
\Crefname{table}{Table}{Tables}
\crefname{table}{Tab.}{Tabs.}

\usepackage[dvipsnames,table]{xcolor}
\usepackage{multirow}
\usepackage[belowskip=-10pt,aboveskip=0pt]{caption}
\usepackage{enumitem}

\newcommand{\vikram}[1]{{\color{green} Vikram: #1}}
\newcommand{\smallsec}[1]{\vspace{0.3pt}\noindent\textbf{#1.}}

\usepackage[accsupp]{axessibility}

\def\cvprPaperID{1118} 
\def\confName{CVPR}
\def\confYear{2023}

\begin{document}

\title{Overlooked Factors in Concept-based Explanations:\\Dataset Choice, Concept Learnability, and Human Capability}

\author{Vikram V. Ramaswamy, Sunnie S. Y. Kim, Ruth Fong, Olga Russakovsky\\
Princeton University\\
{\tt\small \{vr23, suhk, ruthfong, olgarus\}@cs.princeton.edu}
}

\maketitle

\begin{abstract}
Concept-based interpretability methods aim to explain a deep neural network model's components and predictions using a pre-defined set of semantic concepts. These methods evaluate a trained model on a new, ``probe'' dataset and correlate the model's outputs with concepts labeled in that dataset.
Despite their popularity, they suffer from limitations that are not well-understood and articulated in the literature. 
In this work, we identify and analyze three commonly overlooked factors in concept-based explanations.
First, we find that the choice of the probe dataset has a profound impact on the generated explanations.
Our analysis reveals that different probe datasets lead to very different explanations, suggesting that the generated explanations are not generalizable outside the probe dataset.  
Second, we find that concepts in the probe dataset are often harder to learn than the target classes they are used to explain, calling into question the correctness of the explanations.
We argue that only easily learnable concepts should be used in concept-based explanations. 
Finally, while existing methods use hundreds or even thousands of concepts, our human studies reveal a much stricter upper bound of 32 concepts or less, beyond which the explanations are much less practically useful.
We discuss the implications of our findings and provide suggestions for future development of concept-based interpretability methods.
Code for our analysis and user interface can be found at \url{https://github.com/princetonvisualai/OverlookedFactors}

\end{abstract}

\section{Introduction}
\label{sec:intro}

Performance and opacity are often correlated in deep neural networks: the highly parameterized nature of these models that enable them to achieve high task accuracy also reduces their interpretability. 
However, in order to responsibly use and deploy them, especially in high-risk settings such as medical diagnoses, we need these models to be interpretable, i.e., understandable by people.
With the growing recognition of the importance of interpretability, many methods have been proposed in recent years to explain some aspects of neural networks and render them more interpretable 
(see \cite{arrieta2019explainable,fong2020thesis,gilpin2018explaining,rudin2021survey,samek2019book,Zhang2018survey} for surveys).

\begin{figure*}
    \centering
    \includegraphics[width=\textwidth]{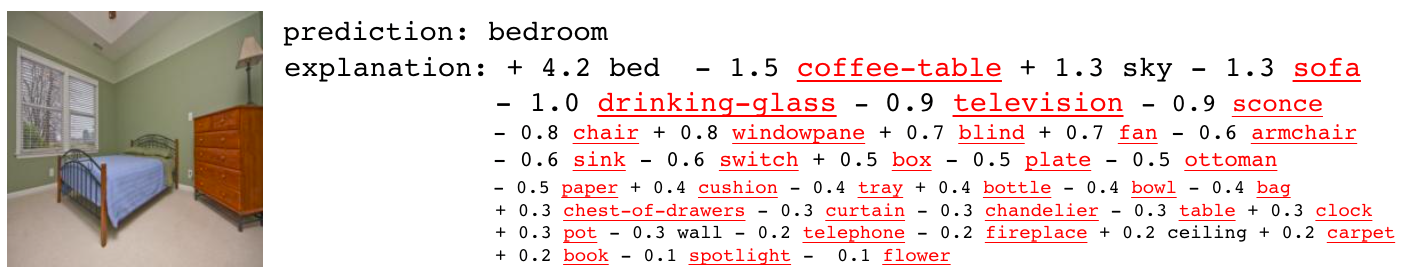}
    \caption{\textbf{Concept-based interpretability methods} explain model components and/or predictions using a pre-defined set of semantic concepts. In this example, a scene classification model's prediction \texttt{bedroom} is explained as a complex linear combination of 37 visual concepts, with the final explanation score calculated based on the presence or absence of these concepts. The coefficients are learned by evaluating the model on a new, ``probe'' dataset, and correlating its predictions with visual concepts labeled in that dataset. However, concept-based explanations can (1) be noisy and heavily dependent on the probe dataset, (2) use concepts that are hard to learn (all concepts in \textcolor{red}{\underline{\texttt{red}}} are harder to learn than the class \texttt{bedroom}) and (3) be overwhelming to people due to the complexity of the explanation.}
    \label{fig:pull_fig}
\end{figure*}

In this work, we dive into \emph{concept-based} interpretability methods for image classification models, which explain model components and/or predictions using a pre-defined set of semantic concepts \cite{bau2017netdissect,fong2018net2vec,kim2018tcav,koh2020conceptbottleneck,zhou2018ibd}.
Given access to a trained model and a set of images labelled with semantic concepts (i.e., a ``probe'' dataset), 
these methods produce explanations with the provided concepts.
See \cref{fig:pull_fig} for an example explanation.

Concept-based methods are a particularly promising approach for bridging the interpretability gap between complex models and human understanding, as they explain model components and predictions with human-interpretable units, i.e., semantic concepts.
Recent work finds that people prefer concept-based explanations over other forms (e.g., heatmap and example-based) because they resemble human reasoning and explanations~\cite{kim2022helpmehelptheai}.
Further, concept-based methods uniquely provide a \textit{global}, high-level understanding of a model, e.g., how it predicts a certain class~\cite{zhou2018ibd,ramaswamy2022elude} and what the model (or some part of it) has learned~\cite{kim2018tcav,bau2017netdissect,fong2018net2vec}.
These insights are difficult to gain from \textit{local} explanation methods that only provide an explanation for a single model prediction, such as saliency maps that highlight relevant regions within an image. 


However, existing research on concept-based interpretability methods focuses heavily on new method development, ignoring important factors such as the probe dataset used to generate explanations or the concepts composing the explanations.
Outside the scope of concept-based methods, there have been several recent works that study the effect of different factors on explanations. These works, however, are either limited to saliency maps~\cite{adebayo2018sanity,kindermans2019reliability,mahendran2016salient,rebuffi2020there} or a general call for transparency, e.g., include more information when releasing an interpretability method~\cite{sokol2020explainability}. 

In this work, we conduct an in-depth study of commonly overlooked factors in concept-based interpretability methods.
Concretely, we analyze four representative methods: NetDissect~\cite{bau2017netdissect}, TCAV~\cite{kim2018tcav}, Concept Bottleneck~\cite{koh2020conceptbottleneck} and IBD~\cite{zhou2018ibd}.
These are a representative and comprehensive set of existing concept-based interpretability methods for computer vision models.
Using multiple probe datasets (ADE20k~\cite{zhou2017ade20k,zhou2019ade20k_ijcv} and Pascal~\cite{Everingham10pascal} for NetDissect, TCAV and IBD; CUB-200-2011~\cite{WahCUB_200_2011} for Concept Bottleneck), we examine the effects of (1) the choice of probe dataset, (2) the concepts used within the explanation, and (3) the complexity of the explanation.
Through our analyses, we learn a number of key insights, which we summarize below:
\vspace{-\topsep}
\begin{itemize}[leftmargin=*,itemsep=2pt]
\item \textbf{The choice of the probe dataset has a profound impact on explanations}. 
We repeatedly find that different probe datasets give rise to different explanations, when explaining the same model with the same interpretability method.
For instance, the prediction of the \texttt{arena/hockey} class is explained with concepts \{\texttt{grandstand}, \texttt{goal}, \texttt{ice-rink}, \texttt{skate-board}\} with one probe dataset, and \{\texttt{plaything}, \texttt{road}\} with another probe dataset.
We highlight that concept-based explanations are not solely determined by the model or the interpretability method.
Hence, probe datasets should be chosen with caution. Specifically, we suggest using probe datasets whose data distribution is similar to that of the dataset the model-being-explained was trained on.
\item \textbf{Concepts used in explanations are frequently harder to learn than the classes they aim to explain}. The choice of concepts used in explanations is dependent on the available concepts in the probe dataset. 
Surprisingly, we find that learning some of these concepts is harder than learning the target classes. For example, in one experiment we find that the target class \texttt{bathroom} is explained using concepts \{\texttt{toilet}, \texttt{shower}, \texttt{countertop}, \texttt{bathtub}, \texttt{screen-door}\}, all of which are harder to learn than \texttt{bathroom}. Moreover, these concepts can be hard for people to identify, limiting the usefulness of these explanations. We argue that learnability is a necessary (albeit not sufficient) condition for the correctness of the explanations, and advocate for future explanations to only use concepts that are  easily learnable.\footnote{Ideally, future methods would also include \emph{causal} rather than purely correlation-based explanations.} 
\item \textbf{Current explanations use hundreds or even thousands of concepts, but human studies reveal a much stricter upper bound.} We conduct human studies with 125 participants recruited from Amazon Mechanical Turk to understand how well people reason with concept-based explanations with varying number of concepts. We find that participants struggle to identify relevant concepts in images as the number of concepts increases (the percentage of concepts recognized per image decreases from $71.7\% \pm 27.7\%$ with 8 concepts to $56.8\% \pm 24.9\%$ for 32 concepts). Moreover, the majority of the participants prefer that the number of concepts be limited to 32. We also find that concept-based explanations offer little to no advantage in predicting model output compared to example-based explanations (the participants' mean accuracy at predicting the model output when given access to explanations with 8 concepts is $64.8\% \pm 23.9\%$ whereas the accuracy when given access to example-based explanations is $60.0\% \pm 30.2\%$). 
\end{itemize}
\vspace{-\topsep}

These findings highlight the importance of vetting intuitions when developing and using interpretability methods. 
We have open-sourced our analysis code and human study user interface to aid with this process in the future: \url{https://github.com/princetonvisualai/OverlookedFactors}.

\section{Related work}
\label{sec:related}

Interpretability methods for computer vision models range from highlighting areas within an image that contribute to a model's prediction (i.e., saliency maps) \cite{chattopadhay2018gradcamplusplus,fong19understanding,Petsiuk2018rise,selvaraju2017gradcam,simonyan2013saliency,zeiler2014visualizing,zhang2016excitation,zhou2016cam} to labelling model components (e.g., neurons) \cite{bau2017netdissect,fong2018net2vec,kim2018tcav,zhou2018ibd}, highlighting concepts that contribute to the model's prediction~\cite{zhou2018ibd,ramaswamy2022elude} and designing models that are interpretable-by-design \cite{brendel2018approximating,chen2018protopnet,koh2020conceptbottleneck,nauta2021prototree}. 
In this work, we focus on concept-based interpretability methods. 
These include post-hoc methods that label a trained model's components and/or predictions~\cite{bau2017netdissect,fong2018net2vec,kim2018tcav,zhou2018ibd,ramaswamy2022elude} and interpretable-by-design methods that use pre-defined concepts~\cite{koh2020conceptbottleneck}.
We focus on methods for image classification models where most interpretability research has been and is being conducted.
Recently, concept-based methods are being developed and used for other types of models (e.g., image similarity models~\cite{plummer2020imgsim}, language models~\cite{yeh2020completeness,bolubasi2021BERT}), however, these are outside the scope of this paper.

Our work is similar in spirit to a growing group of works that propose checks and evaluation protocols to better understand the capabilities and limitations of interpretability methods~\cite{adebayo2018sanity,adebayo2020neurips,adebayo2022iclr,hoffmann2021looks,hooker2019roar,kim2022hive,kindermans2019reliability,margeloiu2021concept,rebuffi2020there,yang2019benchmarking}. Many of these works examine how sensitive post-hoc saliency maps are to different factors such as input perturbations, model weights, or the output class being explained. On the other hand, we conduct an in-depth study of concept-based interpretability methods. Despite their popularity, little is understood about their interpretability and usefulness to human users, or their sensitivity to auxiliary inputs such as the probe dataset. We seek to fill this gap with our work and assist with future development and use of concept-based interpretability methods. To the best of our knowledge, we are the first to investigate the effect of the probe dataset and concepts used for concept-based explanations.
There has been work investigating the effect of explanation complexity on human understanding~\cite{lage2019human}, however, it is limited to decision sets.

We also echo the call for releasing more information when releasing datasets~\cite{gebru2021datasheets}, models~\cite{crisan2022modelcards,mitchell2019model} and interpretability methods~\cite{sokol2020explainability}. More concretely, we suggest that concept-based interpretability method developers to include results from our proposed analyses in their method release, in addition to filling out the explainability fact sheet proposed by Sokol et al.~\cite{sokol2020explainability}, to aid researchers and practitioners to better understand, use, and build on these methods.


\section{Dataset choice: Probe dataset has a profound impact on the explanations}
\label{sec:dataset}

\begin{table*}[t]
    \centering
    
    \resizebox{\textwidth}{!}{%
    \begin{tabular}{lp{8.3cm}p{8.5cm}}
    \toprule
    Scene class & \multicolumn{1}{l}{Top concepts from ADE20k-generated explanations} & \multicolumn{1}{l}{Top concepts from Pascal-generated explanations} \\
    \midrule
    \texttt{arena/hockey} & \textbf{\texttt{grandstand}}, \textbf{\texttt{goal}}, \textbf{\texttt{ice-rink}}, \textbf{\texttt{scoreboard}} & \textbf{\texttt{plaything}}, \textbf{\texttt{road}}\\
    \texttt{auto-showroom} & \texttt{car}, \textbf{\texttt{light}}, \textbf{\texttt{trade-name}}, \textbf{\texttt{floor}}, \textbf{\texttt{wall}} & \texttt{car}, \textbf{\texttt{stage}}, \textbf{\texttt{grandstand}}, \textbf{\texttt{baby-buggy}}, \textbf{\texttt{ground}}\\
    \texttt{bedroom} & \texttt{bed}, \textbf{\texttt{cup}}, \textbf{\texttt{tapestry}}, \textbf{\texttt{lamp}}, \textbf{\texttt{blind}} & \texttt{bed}, \textbf{\texttt{frame}}, \textbf{\texttt{wood}}, \textbf{\texttt{sofa}}, \textbf{\texttt{bedclothes}}\\
    \texttt{bow-window} & \texttt{windowpane}, \textbf{\texttt{seat}}, \textbf{\texttt{cushion}}, \textbf{\texttt{wall}}, \textbf{\texttt{heater}} & \texttt{windowpane}, \textbf{\texttt{tree}}, \textbf{\texttt{shelves}}, \textbf{\texttt{curtain}}, \textbf{\texttt{cup}}\\
    \texttt{conf-room} & \textbf{\texttt{swivel-chair}}, \texttt{table}, \textbf{\texttt{mic}}, \textbf{\texttt{chair}}, \textbf{\texttt{document}} & \textbf{\texttt{bench}}, \textbf{\texttt{napkin}}, \textbf{\texttt{plate}}, \textbf{\texttt{candle}}, \texttt{table}\\
    \texttt{corn-field} & \textbf{\texttt{field}}, \textbf{\texttt{plant}}, \texttt{sky}, \textbf{\texttt{streetlight}} & \textbf{\texttt{tire}}, \texttt{sky}, \textbf{\texttt{dog}}, \textbf{\texttt{water}}, \textbf{\texttt{signboard}}\\
    \texttt{garage/indoor} & \texttt{bicycle}, \textbf{\texttt{brush}}, \textbf{\texttt{car}}, \textbf{\texttt{tank}}, \textbf{\texttt{ladder}} & \texttt{bicycle}, \textbf{\texttt{vending-mach}},  \textbf{\texttt{tire}}, \textbf{\texttt{motorbike}}, \textbf{\texttt{floor}}\\
    \texttt{hardware-store} & \texttt{shelf}, \textbf{\texttt{merchandise}}, \textbf{\texttt{pallet}}, \textbf{\texttt{videos}}, \texttt{box} & \textbf{\texttt{rope}}, \texttt{shelves}, \texttt{box}, \textbf{\texttt{bottle}}, \textbf{\texttt{pole}}\\
    \texttt{legis-chamber} & \textbf{\texttt{seat}}, \textbf{\texttt{chair}}, \textbf{\texttt{pedestal}}, \textbf{\texttt{flag}}, \textbf{\texttt{witness-stand}} & \textbf{\texttt{mic}}, \textbf{\texttt{book}}, \textbf{\texttt{paper}}\\
    \texttt{tree-farm} & \texttt{tree}, \textbf{\texttt{hedge}}, \textbf{\texttt{land}}, \textbf{\texttt{path}}, \textbf{\texttt{pole}} & \texttt{tree}, \textbf{\texttt{tent}}, \textbf{\texttt{sheep}}, \textbf{\texttt{mountain}}, \textbf{\texttt{rock}}\\
  \bottomrule
  \end{tabular}
  }
   \caption{\textbf{Impact of probe dataset on \textit{Baseline} (\cref{sec:dataset}).} We compare \textit{Baseline} explanations generated using ADE20k vs. Pascal. For 10 randomly selected scene classes, we show concepts with the largest coefficients in each explanation. In \textbf{bold} are concepts in one explanation but not the other, e.g., the concept \textbf{\texttt{grandstand}} is important for explaining the \texttt{arena/hockey} scene prediction when using ADE20k, but not when using Pascal. These results show that the probe dataset has a huge impact on the explanations.}
    \label{tab:baseline_dset} 
\end{table*}

\begin{table}[t]
    \centering

    \resizebox{\linewidth}{!}{%
    \begin{tabular}{ccccc}
    \toprule
    Neuron & \multicolumn{2}{c}{ADE20k label \& score} & \multicolumn{2}{c}{Pascal label \& score} \\ 
    \midrule
    9 & \texttt{plant} & 0.082 & \texttt{potted-plant} & 0.194 \\
    181 & \texttt{plant} & 0.068 & \texttt{potted-plant} & 0.140 \\
    318 & \texttt{computer} & 0.079 & \texttt{tv} & 0.251 \\
    386 & \texttt{autobus} & 0.067 & \texttt{bus} & 0.200 \\
    435 & \texttt{runway} & 0.071 & \texttt{\texttt{airplane}} & 0.189 \\
    \midrule
    185 & \texttt{\texttt{chair}} & 0.077 & \texttt{\texttt{horse}} & 0.153 \\
    239 & \texttt{pool-table} & 0.069 & \texttt{\texttt{horse}} & 0.171 \\
    257 & \texttt{tent} & 0.042 & \texttt{bus} & 0.279\\
    384 & \texttt{washer} & 0.043 & \texttt{\texttt{bicycle}} & 0.201 \\
    446 & \texttt{pool-table} & 0.193 & \texttt{tv} & 0.086 \\
    \bottomrule
    \end{tabular}
    }
    \caption{\textbf{Impact of probe dataset on \textit{NetDissect}~\cite{bau2017netdissect} (\cref{sec:dataset}).} 
    We compare NetDissect explanations (concept labels) for 10 neurons of the model-being-explained generated using ADE20k vs. Pascal.
    We find that while some neurons correspond to the same or similar concepts (top half), others correspond to wildly different concepts (bottom half), highlighting the impact of the probe dataset.}
    \label{tab:netdissect_dset}

\end{table}

Concept-based explanations are generated by running a trained model on a ``probe'' dataset (typically not the training dataset) which has concepts labelled within it.
The choice of probe dataset has been almost entirely dictated by which datasets have concept labels.
The most commonly used dataset is the Broden dataset~\cite{bau2017netdissect}.
It contains images from four datasets (ADE20k~\cite{zhou2017ade20k,zhou2019ade20k_ijcv}, Pascal~\cite{Everingham10pascal}, OpenSurfaces~\cite{bell2014surfaces},
Describable Textures Dataset~\cite{Cimpoi2014DTD}) and labels of over 1190 concepts, comprising of object, object parts, color, scene and texture.

In this section, we investigate the effect of the probe dataset by comparing explanations generated using two different subsets of the Broden datset: ADE20k and Pascal. 
We experiment with three different methods for generating concept-based explanations: \textit{Baseline}, \textit{NetDissect}~\cite{bau2017netdissect}, and \textit{TCAV}~\cite{kim2018tcav}, and find that the generated explanations heavily depend on the choice of probe dataset. 
This finding implies that these explanations can only be used for images drawn from the same distribution as the probe dataset. 

\smallsec{Model explained}
Following prior work~\cite{bau2017netdissect,kim2018tcav,zhou2018ibd}, we explain a ResNet18-based~\cite{he2016resnet} scene classification model trained on the Places365 dataset~\cite{zhou2017places}, which predicts one of 365 scene classes given an input image.

\smallsec{Probe datasets}
We use two probe datasets: ADE20k~\cite{zhou2017ade20k,zhou2019ade20k_ijcv} (19733 images, license: BSD 3-Clause) and Pascal~\cite{Everingham10pascal} (10103 images, license: unknown).\footnote{To our best knowledge, most images used don't include personally identifiable information or offensive content. However, some feature people without their consent and might contain identifiable information.} 
They are two different subsets of the Broden dataset~\cite{bau2017netdissect} and are labelled with objects and parts.
We randomly split each dataset into training (60\%), validation (20\%), and test (20\%) sets, using the new training set for learning explanations, validation set for tuning hyperparameters (e.g., learning rate and regularization parameters), and test set for reporting our findings. 

\smallsec{Interpretability methods} We investigate the effect of the probe dataset on three types of concept-based explanations. First, we study a simple \textit{Baseline} method that measures correlations between the model's prediction and concepts, and generates class-level explanations as a linear combination of concepts as in \cref{fig:pull_fig}. Similar to Ramaswamy et al.~\cite{ramaswamy2022elude}, we learn a logistic regression model that matches the model-being-explained's prediction, given access to ground-truth concept labels within the image. We use an \texttt{l1} penalty to prioritize explanations with fewer concepts. Second, we study \textit{NetDissect}~\cite{bau2017netdissect} which identifies neurons within the model-being-explained that are highly activated by certain concepts and generates neuron-level explanations (concept labels).\footnote{We use code provided by the authors: \url{https://github.com/CSAILVision/NetDissect-Lite}.}
Finally, we study \textit{TCAV}~\cite{kim2018tcav} which generates explanations in the form of concept activation vectors, i.e., vectors within the model-being-explained's feature space that correspond to labelled concepts.

\smallsec{Results}
For all three explanation types, we find that using different probe datasets result in very different explanations. 
To begin, we show in \cref{tab:baseline_dset} how \textit{Baseline} explanations differ when using ADE20k vs. Pascal as the probe dataset.
For example, when explaining the \texttt{corn-field} scene prediction, the Pascal-generated explanation highlights \texttt{dog} as important, whereas the ADE20k-generated explanation does not. For the \texttt{legis-chamber} scene, ADE20k highlights \texttt{chair} as important, whereas Pascal does not. 

We observe a similar difference for \textit{NetDissect} (see \cref{tab:netdissect_dset}).
We label 123 neurons separately using ADE20k and Pascal, and find that 60 of them are given very different concept labels (e.g., neuron 239 is labelled \texttt{pool-table} by ADE20k and \texttt{horse} by Pascal).\footnote{It is possible that these neurons are poly-semantic, i.e., neurons that reference multiple concepts, as noted in \cite{fong2018net2vec,olah2020zoom}. 
However, as we explore in the supp. mat., the score for the concept from the other dataset is usually below 0.04, the threshold used in \cite{bau2017netdissect} to identify ``highly activated neurons.'' }
Again, this result highlights the impact of the probe dataset on explanations.

\begin{figure}[t]
    \centering
    \resizebox{0.91\linewidth}{!}{
    \begin{tabular}{cccc}
        \toprule
        \multicolumn{1}{c}{Concept} & \multicolumn{1}{c}{ADE20k AUC} & \multicolumn{1}{c}{Pascal AUC} & \multicolumn{1}{c}{\begin{tabular}[c]{@{}c@{}}Cosine sim\end{tabular}} \\
        \midrule
        \texttt{ceiling} & 96.6 & 93.0 & 0.267 \\ 
        \texttt{box} & 83.0 & 80.1 & 0.086 \\ 
        \texttt{pole} & 89.0 & 79.3 & 0.059 \\ 
        \texttt{bag} & 79.4 & 75.4 & 0.006 \\ 
        \texttt{rock} & 92.6 & 82.8 & -0.024 \\ 
        \midrule
        mean & 92.0 & 88.1 & 0.087 \\
        \bottomrule
    \end{tabular}
    }
    \includegraphics[width=0.8\linewidth]{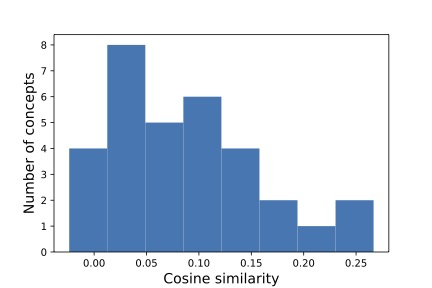}
    \caption{\textbf{Impact of probe dataset on \textit{TCAV}~\cite{kim2018tcav} (\cref{sec:dataset}).} 
    We compare TCAV concept activation vectors learned using ADE20k vs. Pascal.
    \emph{(Top)} For 5 concepts randomly selected out of 32, we show their learnability in each dataset (AUC) and cosine similarity between the two vectors. While these concepts can be learned reasonably well (AUCs are high), their learned activation vectors have low similarity (Cosine sim is low).
    \emph{(Bottom)} The histogram of cosine similarity scores for all 32 concepts again shows that the two activation vectors for the same concept are not very similar.}
    \label{fig:tcav_hist}
    \vspace{-0.25cm}
\end{figure}

Similarly, \textit{TCAV} concept activation vectors learned using ADE20k vs. Pascal are different, i.e., they have low cosine similarity (see \cref{fig:tcav_hist}).
We compute concept activation vectors for 32 concepts which have a base rate of over 1\% in both datasets combined, 
then calculate the cosine similarity of each concept vector. We also compute the ROC AUC for each concept vector to measure how well the concept vector corresponds to the concept. 
We find that the similarity is low (0.078 on average), even though the selected concepts were those that can be learned reasonably well (mean ROC AUC for these concepts is over 85\%). 
We suspect that the explanations are radically different due to differences in the probe dataset distribution.
For instance, some concepts have very different base rates in the two datasets: \texttt{dog} has a base rate of 12.0\% in Pascal but 0.5\% in ADE20k; \texttt{chair} has a base rate of 16.7\% in ADE20k but 13.5\% in Pascal. We present more analyses in the supp. mat.

\section{Concept learnability: Concepts used are less learnable than target classes}
\label{sec:corr}

In \cref{sec:dataset}, we investigated how the choice of the probe dataset influences the generated explanations. In this section, we investigate the individual concepts used within explanations.
An implicit assumption made in concept-based interpretability methods is that the concepts used in explanations are easier to learn than the target classes being explained. 
For instance, when explaining the class \texttt{bedroom} with the concept \texttt{bed}, we are assuming (and hoping) that the model first learns the concept \texttt{bed}, then uses this concept and others to predict the class \texttt{bedroom}.
However, if \texttt{bed} is harder to learn than \texttt{bedroom}, this would not be the case. 
This assumption also aligns with works that argue that ``simpler'' concepts (i.e., edges and textures) are learned in early layers and ``complex'' concepts (i.e., parts and objects) are learned in later layers~\cite{bau2017netdissect,fong2018net2vec}.

We thus investigate the learnability of concepts used by different explanation methods. 
Somewhat surprisingly, we find that the concepts used are frequently harder to learn than the target classes, raising concerns about the correctness of concept-based explanations.

\begin{figure}[t]
\begin{center}
    \begin{minipage}{0.45\textwidth}
        \centering
        \includegraphics[width=0.9\textwidth]{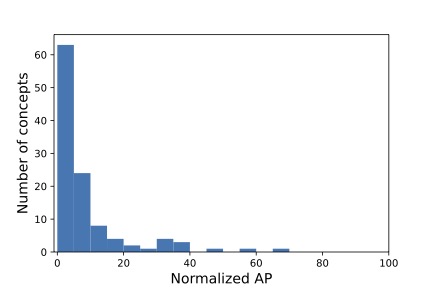}
    \end{minipage}
    \begin{minipage}{0.45\textwidth}
        \centering
        \includegraphics[width=0.9\textwidth]{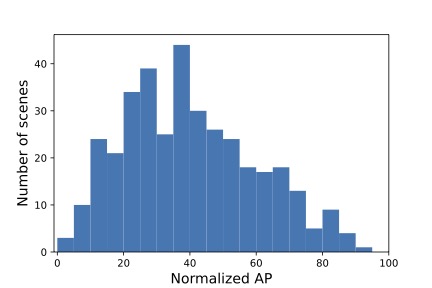}
    \end{minipage}
    \caption{\textbf{Overall comparison of concept vs. class learnability (\cref{sec:corr}).} We compare the learnability, quantified as normalized AP of concept/class predictors, of Broden concepts (\emph{top}) vs. Places365 scene classes (\emph{below}). 
    Overall, the concepts have much lower normalized AP (i.e., are harder to learn) than the classes.
    }
    \label{fig:attr_vs_scene}
\end{center}
\vspace{-0.25cm}
\end{figure}

\smallsec{Setup} 
To compare the learnability of concepts vs. classes, we learn models for the concepts (the learnability of the classes is already known from the model-being-explained). Concretely, we extract features for the probe dataset using an ImageNet~\cite{russakovsky2015imagenet}-pretrained ResNet18~\cite{he2016resnet} model and train a linear model using sklearn's~\cite{scikit-learn} \texttt{LogisticRegression} to predict concepts from the ResNet18 features.\footnote{We also tried using features from a Places365 pretrained model and did not find a significant difference.} 
We do so for the two most commonly used probe datasets: Broden~\cite{bau2017netdissect} and CUB-200-2011~\cite{WahCUB_200_2011}. Broden concepts are frequently used to explain Places365 classes (as done in NetDissect~\cite{bau2017netdissect}, Net2Vec~\cite{fong2018net2vec}, IBD~\cite{zhou2018ibd}, and ELUDE~\cite{ramaswamy2022elude}), while CUB concepts are used to explain the CUB target classes (as done in Concept Bottleneck~\cite{koh2020conceptbottleneck} and ELUDE~\cite{ramaswamy2022elude}). 

\smallsec{Evaluation} 
We evaluate learnability with normalized average precision (AP)~\cite{hoiem2012error}. We choose normalized AP for two reasons: first, to avoid having to set a threshold and second, to fairly compare concepts and scenes that have very different base rates. In our experiments, we set the base rate to be that of the classes: $\frac{1}{365}$ when comparing Broden concepts vs. Places365 classes and $\frac{1}{200}$ when comparing CUB concepts vs. CUB classes. 

\smallsec{Results}
In both settings, we find that the concepts are much harder to learn than the target classes.
The median normalized AP for Broden concepts is 7.6\%, much lower than 37.5\% of Places365 classes. Similarly, the median normalized AP for CUB concepts is 2.3\%, much lower than 65.9\% of CUB classes.
Histograms of normalized APs are shown in \cref{fig:attr_vs_scene} (Broden/Places365) and the supp. mat. (CUB).

However, is it possible that each class is explained by concepts that are more learnable than the class?
Our investigation with IBD~\cite{zhou2018ibd} explanations suggests this is not the case.
IBD greedily learns a basis of concept vectors, as well as a residual vector, and decomposes each model prediction into a linear combination of the basis and residual vectors.\footnote{We use code provided by the authors: \url{https://github.com/CSAILVision/IBD}.} 
For 10 randomly chosen scene classes, we compare the normalized AP of the scene class vs. 5 concepts with the highest coefficients (i.e., 5 concepts that are the most important for explaining the prediction).
See \cref{tab:ibd_attr_vs_scenes} for the results. 
We find that all 10 scene classes are explained with at least one concept that is harder to learn than the class.
For some classes (e.g., \texttt{bathroom}, \texttt{kitchen}), all concepts used in the explanation are harder to learn than the class.

\begin{table*}[ht]
    \centering
    \resizebox{0.87\textwidth}{!}{%
        \begin{tabular}{@{}c|ccccc@{}}
\toprule
\large{Scene class} & \multicolumn{5}{c}{\large{Concepts}} \\
\midrule
\texttt{arena/perform} & \textcolor{NavyBlue}{\textbf{\texttt{tennis court}}} & \textcolor{NavyBlue}{\textbf{\texttt{grandstand}}} & \textcolor{NavyBlue}{\textbf{\texttt{ice rink}}} & \textcolor{red}{\textit{\texttt{valley}}} & \textcolor{red}{\textit{\texttt{stage}}} \\
38.8 & \textcolor{NavyBlue}{\textbf{\texttt{74.0}}} & \textcolor{NavyBlue}{\textbf{\texttt{44.4}}} & \textcolor{NavyBlue}{\textbf{\texttt{40.7}}} & \textcolor{red}{\textit{\texttt{19.0}}} &  \textcolor{red}{\textit{\texttt{11.9}}}\\
\arrayrulecolor{gray}\midrule
\texttt{art-gallery} & \textcolor{NavyBlue}{\textbf{\texttt{binder}}} & \textcolor{red}{\textit{\texttt{drawing}}} & \textcolor{red}{\textit{\texttt{painting}}} & \textcolor{red}{\textit{\texttt{frame}}} & \textcolor{red}{\textit{\texttt{sculpture}}} \\
27.4 & \textcolor{NavyBlue}{\textbf{\texttt{42.6}}} & \textcolor{red}{\textit{\texttt{10.8}}} & \textcolor{red}{\textit{\texttt{10.5}}} & \textcolor{red}{\textit{\texttt{2.5}}} & \textcolor{red}{\textit{\texttt{0.7}}} \\
\midrule
\texttt{bathroom} & \textcolor{red}{\textit{\texttt{toilet}}} & \textcolor{red}{\textit{\texttt{shower}}} & \textcolor{red}{\textit{\texttt{countertop}}} &  \textcolor{red}{\textit{\texttt{bathtub}}} & \textcolor{red}{\textit{\texttt{screen door}}}\\
43.3 & \textcolor{red}{\textit{\texttt{39.9}}} & \textcolor{red}{\textit{\texttt{18.8}}} & \textcolor{red}{\textit{\texttt{12.6}}} & \textcolor{red}{\textit{\texttt{11.1}}} & \textcolor{red}{\textit{\texttt{9.6}}}\\
\midrule
\texttt{kasbah} & \textcolor{NavyBlue}{\textbf{\texttt{ruins}}} & \textcolor{red}{\textit{\texttt{desert}}} & \textcolor{red}{\textit{\texttt{arch}}} & \textcolor{red}{\textit{\texttt{dirt track}}} & \textcolor{red}{\textit{\texttt{bottle rack}}}\\
50.2 & \textcolor{NavyBlue}{\textbf{\texttt{64.3}}} & \textcolor{red}{\textit{\texttt{17.3}}} & \textcolor{red}{\textit{\texttt{16.2}}} & \textcolor{red}{\textit{\texttt{8.9}}} & \textcolor{red}{\textit{\texttt{4.2}}}\\
\midrule
\texttt{kitchen} & \textcolor{red}{\textit{\texttt{work surface}}} & \textcolor{red}{\textit{\texttt{stove}}} & \textcolor{red}{\textit{\texttt{cabinet}}} & \textcolor{red}{\textit{\texttt{refrigerator}}} & \textcolor{red}{\textit{\texttt{doorframe}}}\\
33.9 & \textcolor{red}{\textit{\texttt{24.8}}} & \textcolor{red}{\textit{\texttt{18.2}}} & \textcolor{red}{\textit{\texttt{10.3}}} & \textcolor{red}{\textit{\texttt{8.8}}} & \textcolor{red}{\textit{\texttt{2.8}}}\\
\midrule
\texttt{lock-chamber} & \textcolor{NavyBlue}{\textbf{\texttt{water wheel}}} & \textcolor{NavyBlue}{\textbf{\texttt{dam}}} & \textcolor{red}{\textit{\texttt{boat}}} &  \textcolor{red}{\textit{\texttt{embankment}}} & \textcolor{red}{\textit{\texttt{footbridge}}} \\
36.5 & \textcolor{NavyBlue}{\textbf{\texttt{47.4}}} &  \textcolor{NavyBlue}{\textbf{\texttt{43.7}}} & \textcolor{red}{\textit{\texttt{16.1}}} & \textcolor{red}{\textit{\texttt{4.8}}} &  \textcolor{red}{\textit{\texttt{4.1}}}\\
\midrule
\texttt{pasture} & \textcolor{NavyBlue}{\textbf{\texttt{cow}}} & \textcolor{NavyBlue}{\textbf{\texttt{leaf}}} & \textcolor{red}{\textit{\texttt{valley}}} & \textcolor{red}{\textit{\texttt{field}}}   & \textcolor{red}{\textit{\texttt{slope}}}\\
19.2 & \textcolor{NavyBlue}{\textbf{\texttt{63.7}}} & \textcolor{NavyBlue}{\textbf{\texttt{21.1}}} & \textcolor{red}{\textit{\texttt{19.0}}}  & \textcolor{red}{\textit{\texttt{6.8}}} & \textcolor{red}{\textit{\texttt{4.1}}}\\
\midrule
\texttt{physics-lab} & \textcolor{NavyBlue}{\textbf{\texttt{computer}}} & \textcolor{red}{\textit{\texttt{machine}}} & \textcolor{red}{\textit{\texttt{monitor-device}}} & \textcolor{red}{\textit{\texttt{bicycle}}} & \textcolor{red}{\textit{\texttt{sewing-machine}}}\\
17.1 & \textcolor{NavyBlue}{\textbf{\texttt{25.4}}} & \textcolor{red}{\textit{\texttt{4.5}}} & \textcolor{red}{\textit{\texttt{3.3}}} & \textcolor{red}{\textit{\texttt{1.7}}} & \textcolor{red}{\textit{\texttt{1.5}}}\\
\midrule
\texttt{store/indoor}  & \textcolor{NavyBlue}{\textbf{\texttt{shanties}}} & \textcolor{red}{\textit{\texttt{patty}}} & \textcolor{red}{\textit{\texttt{bookcase}}} & \textcolor{red}{\textit{\texttt{shelf}}} & \textcolor{red}{\textit{\texttt{cup}}}\\
20.4 & \textcolor{NavyBlue}{\textbf{\texttt{72.5}}} & \textcolor{red}{\textit{\texttt{18.5}}} & \textcolor{red}{\textit{\texttt{13.5}}} & \textcolor{red}{\textit{\texttt{4.2}}} & \textcolor{red}{\textit{\texttt{1.3}}}\\
\midrule
\texttt{water-park} & \textcolor{NavyBlue}{\textbf{\texttt{roller coaster}}} & \textcolor{NavyBlue}{\textbf{\texttt{hot tub}}} & \textcolor{NavyBlue}{\textbf{\texttt{playground}}} & \textcolor{red}{\textit{\texttt{ride}}}  & \textcolor{red}{\textit{\texttt{swimming pool}}} \\
38.3 & \textcolor{NavyBlue}{\textbf{\texttt{73.0}}} & \textcolor{NavyBlue}{\textbf{\texttt{59.1}}} & \textcolor{NavyBlue}{\textbf{\texttt{44.9}}} & \textcolor{red}{\textit{\texttt{38.0}}} & \textcolor{red}{\textit{\texttt{36.7}}} \\
\arrayrulecolor{black}\bottomrule
\end{tabular}
    }
    \caption{\textbf{Class-level comparison of concept vs. class learnability (\cref{sec:corr}).} We report normalized AP scores ($\uparrow$ indicates high learnability) for 10 randomly chosen scene classes, along with 5 concepts with the highest IBD explanation coefficients for each. Concepts whose normalized AP scores are lower than the scene class are shown in \textcolor{red}{\textit{\texttt{red}}}, whereas concepts with higher scores are shown in \textcolor{NavyBlue}{\textbf{\texttt{blue}}}. All scenes are explained by at least one concept with a lower normalized AP. Some scenes are only explained by concepts with lower normalized AP. }
    \label{tab:ibd_attr_vs_scenes}
\end{table*}


Our experiments show that a significant fraction of the concepts used by existing concept-based interpretability methods are harder to learn than the target classes, issuing a wake-up call to the field.
In the following section, we show that these concepts can also be hard for people to identify. 

\section{Human capability: Human studies reveal an upper bound of 32 concepts}
\label{sec:human_exp}

Existing concept-based explanations use a large number of concepts:
NetDissect~\cite{bau2017netdissect} and Net2Vec~\cite{fong2018net2vec} use all 1197 concepts labelled within the Broden~\cite{bau2017netdissect} dataset; IBD~\cite{zhou2018ibd} uses Broden object and art concepts with at least 10 examples (660 concepts); and Concept Bottleneck~\cite{koh2020conceptbottleneck} uses all concepts that are predominantly present for at least 10 classes from CUB~\cite{WahCUB_200_2011} (112 concepts). 
However, can people actually reason with these many concepts?


In this section, we study this important yet overlooked aspect of concept-based explanations: \textit{explanation complexity} and how it relates to human capability and preference. Specifically, we investigate: 
(1) How well do people recognize concepts in images? (2) How do the (concept recognition) task performance and time change as the number of concepts vary? (3) How well do people predict the model output for a new image using explanations? (4) How do people trade off simplicity and correctness of concept-based explanations?
To answer these questions, we design and conduct a human study.
We describe the study design in \cref{sec:human_studydesign} and report findings in \cref{sec:human_findings}.

\subsection{Human study design}
\label{sec:human_studydesign}
We build on the study design and user interface (UI) of HIVE~\cite{kim2022hive}, and design a two-part study to understand how understandable and useful concept-based explanations are to human users with potentially limited knowledge about machine learning . 
To the best of our knowledge, we are the first to investigate such properties of concept-based explanations for computer vision models.\footnote{We note that there are works examining complexity of explanations for other types of models, for example, Lage et al.~\cite{lage2019human} investigate complexity of explanations over decision sets, Bolubasi et al.~\cite{bolubasi2021BERT} investigate this for concept-based explanations for language models.}

\begin{figure*}
    \centering
    \includegraphics[width=\textwidth]{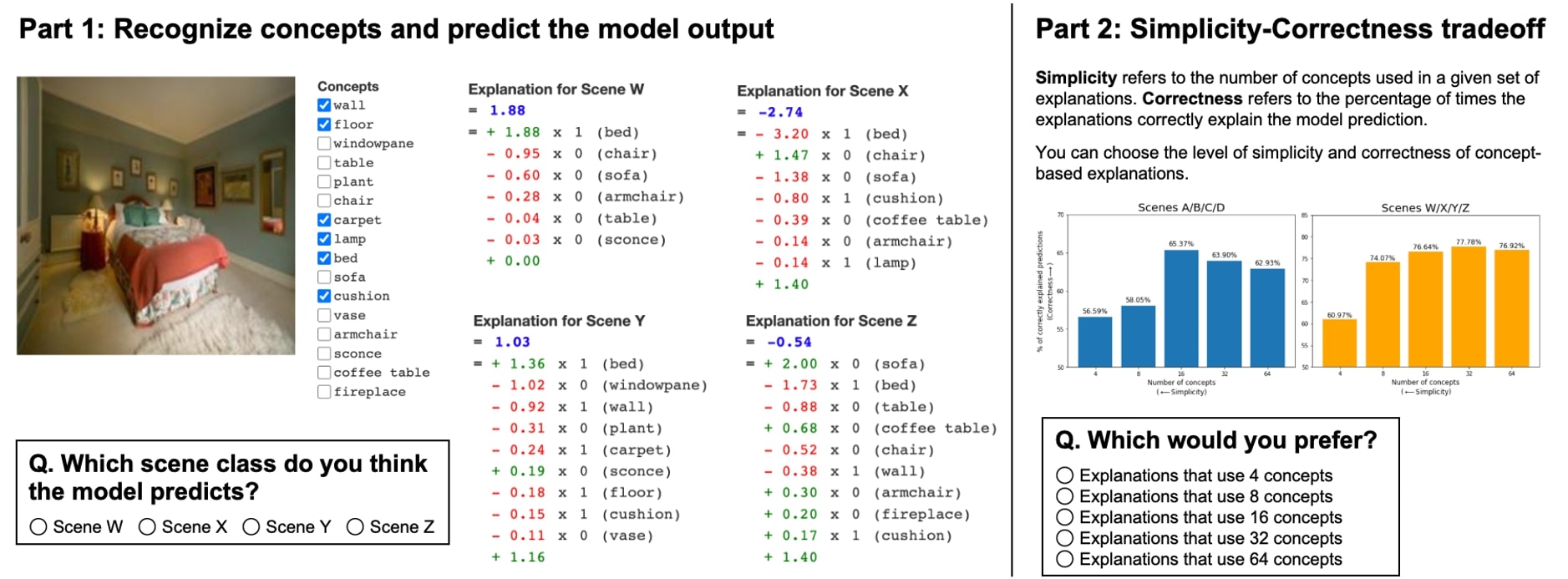}
    \caption{
    \textbf{Human study UI (\cref{sec:human_exp}).}
    We show a simplified version of the UI we developed for our human studies.
    In Part 1, we ask participants to guess the model's prediction for a given image by recognizing concepts and using the provided explanations.
    In Part 2, we show participants explanations with different levels of simplicity and correctness, then ask which one they prefer the most.
    }
    \label{fig:ui}
\end{figure*}

\smallsec{Part 1: Recognize concepts and predict the model output}
First, we present participants with an image and a set of concepts and ask them to identify whether each concept is present or absent in the image.
We also show explanations for 4 classes whose scores are calculated real-time based on the concepts selected. As a final question, we ask participants to select the class they think the model predicts for the given image. See~\cref{fig:ui} (\emph{left}) for the study UI.

To ensure that the task is doable and is only affected by explanation complexity (number of concepts used) and not the complexity of the model and its original prediction task (e.g., 365 scenes classification), we generate explanations for only 4 classes and ask participants to identify which of the 4 classes corresponds to the model's prediction. We only show images where the model output matches the explanation output (i.e., the model predicts the class with the highest explanation score, calculated with ground-truth concept labels), since our goal is to understand how people reason with concept-based explanations with varying complexity.

\smallsec{Part 2: Choose the ideal tradeoff between simplicity and correctness} 
Next, we ask participants to reason about two properties of concept-based explanations: \textit{simplicity}, i.e., the number of concepts used in a given set of explanations, and \textit{correctness}, i.e., the percentage of model predictions correctly explained by explanations, which is the percentage of times the model output class has the highest explanation score. See~\cref{fig:ui} \emph{(right)} for the study UI.
We convey the notion of a simplicity-correctness tradeoff through bar plots that show the correctness of explanations of varying simplicity/complexity (4, 8, 16, 32, 64 concepts). 
We then ask participants to choose the explanation they prefer the most and provide a short justification for their choice.

\smallsec{Full study design and experimental details}
In summary, our study consists of the following steps.
For each participant, we introduce the study, receive informed consent for participation in the study, and collect information about their demographic (optional) and machine learning experience.
We then introduce concept-based explanations in simple terms, and show a preview of the concept recognition and model output prediction task in Part 1. The participant then completes the task for 10 images. 
In Part 2, the participant indicates their preference for explanation complexity, given simplicity and correctness information.
There are no foreseeable risks in participation in the study, and our study design was approved by our institution's IRB.

Using this study design, we investigate explanations that take the form of a linear combination of concepts (e.g., Baseline, IBD~\cite{zhou2018ibd}, Concept Bottleneck~\cite{koh2020conceptbottleneck}). Explanations are generated using the \textit{Baseline} method, which is a logistic regression model trained to predict the model's output using concepts (see Sec.~\ref{sec:dataset} for details). Note that we are evaluating the form of explanation (linear combination of concepts) rather than a specific explanation method. The choice of the method does not impact the task. 

Specifically, we compare four types of explanations: concept-based explanations that use (1) 8 concepts, (2) 16 concepts, (3) 32 concepts, and (4) example-based explanations that consist of 10 example images for which the model predicts a certain class.
We include (4) as a method that doesn't use concepts. 
In Jeyakumar et al.~\cite{jeyakumar2020explain}, this type of explanation is shown to be preferred over saliency-type explanations for image classification; here, we compare this to concept-based explanations. 

For a fair comparison, all four are evaluated on the same set of images.
In short, we conduct a between-group study with 125 participants recruited through Amazon Mechanical Turk.
Participants were compensated based on the state-level minimum wage of \$12/hr. In total, $\sim$\$800 was spent on running human studies.
See supp. mat. for more details.

\subsection{Key findings from the human studies}
\label{sec:human_findings}

\smallsec{When presented with more concepts, participants spend more time but are worse at recognizing concepts}
The median time participants spend on each image is 17.4 sec. for 8 concept-, 27.5 sec. for 16 concept-, and 46.2 sec. for 32 concept-explanations.
This is expected, since participants are asked to make a judgment for each and every concept.
When given example-based explanations with no such task, participants spend only 11.6 seconds on each image.
Interestingly, the concept recognition performance, reported in terms of mean recall (i.e., the percentage of concepts in the image that are recognized) and standard deviation, decreases from 71.7\% $\pm$ 27.7\% (8 concepts) to 61.0\% $\pm$ 28.5\% (16 concepts) to 56.8\% $\pm$ 24.9\% (32 concepts). While these numbers are far from perfect recall (100\%), participants are better at judging whether concepts are present when shown fewer number of concepts.

\smallsec{Concept-based explanations offer little to no advantage in model output prediction over example-based explanations}
Indeed, we see that the participants' errors in concept recognition result in an incorrect class having the highest explanation score. 
When predicting the model output as the class with the highest explanation score, calculated based on the participants' concept selections,
the mean accuracy and standard deviation of model output prediction are 64.8\% $\pm$ 23.9\% (8 concepts), 63.2\% $\pm$ 26.9\% (16 concepts), 63.6\% $\pm$ 22.2\% (32 concepts).
These are barely higher than 60.0\% $\pm$ 30.2\% of example-based explanations, which are simpler and require less time to complete the task. 

\smallsec{The majority of participants prefer explanations with 8, 16, or 32 concepts}
When given options of explanations that use 4, 8, 16, 32, or 64 concepts, 82\% of participants prefer explanations with 8, 16, or 32 concepts (28\%, 33\%, 21\% respectively). 
Only 6\% prefer those with 64 concepts, suggesting that existing explanations that use hundreds or even thousands of concepts do not cater to human preferences.
In the written responses, many favored having fewer concepts (e.g., ``the lesser, the better") and expressed concerns against having too many (e.g., ``I think 32 is a lot, but 16 is an adequate enough number that it could still predict well..."). In making the tradeoff, some valued correctness above all else (e.g., ``Out of all the options, 32 is the most correct"), while others reasoned about marginal benefits (e.g., ``I would prefer explanations that use 16 concepts because it seems that the difference in percentage of correctness is much closer and less, than other levels of concepts").
Overall, we find that participants actively reason about both simplicity and correctness of explanations. 



\section{Discussion}


Our analyses yield immediate suggestions for improving the quality and usability of concept-based explanations.
First, we suggest choosing a probe dataset whose distribution is similar to that of the dataset the model was trained on. Second, we suggest only using concepts that are more learnable than the target classes.
Third, we suggest limiting the number of concepts used within an explanation to under 32, so that explanations are not overwhelming to people.

The final suggestion is easy to implement.
However, the first two are easier said than done, since the number of available probe datasets (i.e., large-scale datasets with concept labels) is minimal, forcing researchers to use the Broden dataset~\cite{bau2017netdissect} or the CUB dataset~\cite{WahCUB_200_2011}.
Hence, we argue creating diverse and high-quality probe datasets is of upmost importance in researching concept-based explanations.

Another concern is that these methods do highlight hard-to-learn concepts when given access to them, suggesting that they sometimes learn correlations rather than causations.
Methods by Goyal et al.~\cite{goyal2019counterfactual}, which output patches within the image that need to be changed for the model's prediction to change, or Fong et al.~\cite{fong19understanding}, which find regions within the image that maximally contribute to the model's prediction, are more in line with capturing causal relationships. However, these only produce \textit{local} explanations, i.e., explanations of a single model prediction, and not class-level \textit{global} explanations.
One approach to capturing causal relationships is to generate counterfactual images with or without certain concepts using generative models~\cite{ramaswamy2021debiasing} and observe changes in model predictions.

\section{Limitations and future work}

Our findings come with a few caveats.
First, due to the lack of available probe datasets, we tested each concept-based interpretability method in a single setting. That is, we tested NetDissect~\cite{bau2017netdissect}, TCAV~\cite{kim2018tcav} and IBD~\cite{zhou2018ibd} on a scene classifier trained on the Places365 dataset~\cite{zhou2017places}, and Concept Bottleneck~\cite{koh2020conceptbottleneck} on the CUB dataset~\cite{WahCUB_200_2011}.
We plan to expand our analyses as more probe datasets become available.
Second, all participants in our human studies were recruited from Amazon Mechanical Turk. This means that our participants represent a population with limited ML background: the self-reported ML experience was 2.5 $\pm$ 1.0 (on a scale of 1 to 5), which is between ``2: have heard about...'' and ``3: know the basics...'' We believe Part 1 results of our human studies (described in Sec.~\ref{sec:human_studydesign}) will not vary with participants’ ML expertise or role in the ML pipeline, as we are only asking participants to identify concepts in images. However, Part 2 results may vary (e.g., developers debugging a ML model may be more willing to trade off explanation simplicity for correctness than lay end-users). 
Investigating differences in perceptions and uses of concept-based explanations, is an important direction for future research.

\section{Conclusion}

In this work, we examined implicit assumptions made in concept-based interpretability methods along three axes: the choice of the probe datasets, the learnability of the used concepts, and the complexity of explanations.
We found that the choice of the probe dataset profoundly influences the generated explanations, implying that these explanations can only be used for images from the probe dataset distribution.
We also found that a significant fraction of the concepts used within explanations are harder for a model to learn than the target classes they aim to explain. 
Finally, we found that people struggle to identify concepts in images when given too many concepts, and that explanations with less than 32 concepts are preferred.
We hope our proposed analyses and findings lead to more careful use and development of concept-based explanations.

\smallsec{Acknowledgements} 
We foremost thank our participants for taking the time to participate in our study. We also thank the authors of~\cite{bau2017netdissect,zhou2018ibd,koh2020conceptbottleneck,kim2022hive} for open-sourcing their code and/or models. Finally, we thank the anonymous reviewers and the Princeton Visual AI Lab members (especially Nicole Meister) who provided helpful feedback on our work.
This material is based upon work partially supported by the National Science Foundation under Grants No. 1763642, 2145198 and 2112562. Any opinions, findings, and conclusions or recommendations expressed in this material are those of the author(s) and do not necessarily reflect the views of the National Science Foundation. We also acknowledge support from the Princeton SEAS Howard B. Wentz, Jr. Junior Faculty Award to OR, Princeton SEAS Project X Fund to RF and OR, Open Philanthropy Grant to RF, and NSF Graduate Research Fellowship to SK. 
{\small
\bibliographystyle{ieee_fullname}
\bibliography{references}
}

\section*{Appendix}

\appendix

In this supplementary document, we provide additional
details on some sections of the main paper.

\begin{itemize}[leftmargin=0pt, itemsep=1pt, topsep=1pt]
\item[] \textbf{Section \ref{sec:supp-exp-details}} We provide more information regarding our experimental setup over all experiments. 

\item[] \textbf{Section \ref{sec:supp-dataset}} We provide additional results from our experiments regarding probe dataset choice from Section 3 of the main paper. 

\item[] \textbf{Section \ref{sec:supp-corr}} We provide additional results from our experiments regarding concept choice from Section 4 of the main paper. 

\item[] \textbf{Section \ref{sec:supp-humanstudy}:} We supplement Section 5 of the main paper and provide more information about our human studies.
\item[] \textbf{Section \ref{sec:supp-ui}:} We supplement Section 5 of the main paper and show snapshots of our full user interface.
\end{itemize}

\section{Experimental details}
\label{sec:supp-exp-details}
Here we provide additional experimental details regarding all our setups, as well as the computational power we needed. 

\smallsec{TCAV} Using the features extracted from the penultimate layer of the ResNet18-based~\cite{he2016resnet} model trained on the Places365 dataset~\cite{zhou2017places}, we use scikit-learns's~\cite{scikit-learn} \texttt{LogisticRegression} models to predict the ground truth attributes in each case. We use the \texttt{liblinear} solver, with an \texttt{l2} penalty, and pick the regularization weight as a hyperparameter, based on the performance (ROC AUC) on a validation set.

\smallsec{Baseline}
Given the ground-truth labelled concepts for an image, this explanation attempts to predict the blackbox model's output on the image.  We use scikit-learn's~\cite{scikit-learn} \texttt{LogisticRegression} model with a \texttt{liblinear} solver, and an \texttt{l1} penalty, to prioritize learning simpler explanations. For the experiment reported in Section 3 of the main paper, we pick the regularization weight as a hyperparamater, choosing the weight with the best performance on a validation set. When generating explanations of different complexities for our human studies, we vary the regularization parameter, picking explanations that use a total of 4, 8, 16, 32, or 64 concepts. 

\smallsec{Learning concepts}
We computed features for all images from ADE20k~\cite{zhou2017ade20k,zhou2019ade20k_ijcv} using the penultimate layer of a ResNet18~\cite{he2016resnet} model trained on  Imagenet~\cite{russakovsky2015imagenet}. We then learned a linear model for all  concepts that had over 10 positive samples within the dataset, using the \texttt{LogisticRegression} model from scikit-learn~\cite{scikit-learn}. Similar to other models, we use a \texttt{liblinear} solver, with an \texttt{l2} penalty, choosing the regularization weight based on performance (ROC AUC) on a validation set. As mentioned, we report the normalized AP~\cite{hoiem2012error} to be able to compare across concepts and target classes with varying base rates. 

\smallsec{Run times} Computing each of the linear models used less than 2 min on a CPU. Computing features using a ResNet18~\cite{he2016resnet} model trained on either Places365~\cite{zhou2017places} or Imagenet~\cite{russakovsky2015imagenet} for the ADE20k~\cite{zhou2017ade20k,zhou2019ade20k_ijcv} and Pascal~\cite{Everingham10pascal} datasets took less than 15 min using a NVIDIA GTX 2080 GPU.

\section{Probe dataset choice: more details}
\label{sec:supp-dataset}
In our first claim, we show that the choice of probe dataset can have a significant impact on the explanation output for concept-based explanations. We give more details from our experiments for this claim within this section. 

\subsection{Varying the probe dataset}
Here we provide the full results from section 3.1 in the main text, where we compute concept-based explanations using 2 different methods (NetDissect~\cite{bau2017netdissect} and TCAV~\cite{kim2018tcav}) when using either ADE20k or Pascal as probe datasets. 

\smallsec{NetDissect} Table~\ref{tab:netdissect_supp} contains the label generated for all neurons that are strongly activated when using either ADE20k~\cite{zhou2017ade20k,zhou2019ade20k_ijcv} or Pascal~\cite{Everingham10pascal} as the probe dataset. A majority of neurons (69/123) correspond to very different concepts. 

\begin{table*}[t]
    \centering
    
    \resizebox{\textwidth}{!}{%
    \begin{tabular}{ccccc|ccccc}
\toprule
    Neuron & ADE20k label & ADE20k score & Pascal label & Pascal score & Neuron & ADE20k label & ADE20k score & Pascal label & Pascal score \\
\midrule
\textcolor{red}{1} & \textcolor{red}{\textit{\texttt{counter}}} & \textcolor{red}{0.059} & \textcolor{red}{\textit{\texttt{bottle}}} & \textcolor{red}{0.049} & 3 & \texttt{sea} & 0.067 & \texttt{water} & 0.065\\ 
\textcolor{red}{4} & \textcolor{red}{\textit{\texttt{seat}}} & \textcolor{red}{0.064} & \textcolor{red}{\textit{\texttt{tvmonitor}}} & \textcolor{red}{0.074} & 8 & \texttt{vineyard} & 0.048 & \texttt{plant} & 0.043\\ 
9 & \texttt{plant} & 0.082 & \texttt{pottedplant} & 0.194 & \textcolor{red}{22} & \textcolor{red}{\textit{\texttt{bookcase}}} & \textcolor{red}{0.07} & \textcolor{red}{\textit{\texttt{bus}}} & \textcolor{red}{0.048}\\ 
30 & \texttt{house} & 0.094 & \texttt{building} & 0.043 & 37 & \texttt{boat} & 0.043 & \texttt{boat} & 0.213\\ 
43 & \texttt{bed} & 0.151 & \texttt{bed} & 0.075 & \textcolor{red}{47} & \textcolor{red}{\textit{\texttt{pool table}}} & \textcolor{red}{0.135} & \textcolor{red}{\textit{\texttt{airplane}}} & \textcolor{red}{0.079}\\ 
60 & \texttt{plane} & 0.052 & \texttt{airplane} & 0.168 & \textcolor{red}{63} & \textcolor{red}{\textit{\texttt{field}}} & \textcolor{red}{0.053} & \textcolor{red}{\textit{\texttt{muzzle}}} & \textcolor{red}{0.042}\\ 
69 & \texttt{person} & 0.047 & \texttt{hair} & 0.086 & \textcolor{red}{73} & \textcolor{red}{\textit{\texttt{water}}} & \textcolor{red}{0.041} & \textcolor{red}{\textit{\texttt{bird}}} & \textcolor{red}{0.080}\\ 
79 & \texttt{plant} & 0.064 & \texttt{pottedplant} & 0.064 & 90 & \texttt{mountain} & 0.071 & \texttt{mountain} & 0.066\\ 
\textcolor{red}{102} & \textcolor{red}{\textit{\texttt{bathtub}}} & \textcolor{red}{0.040} & \textcolor{red}{\textit{\texttt{cat}}} & \textcolor{red}{0.055} & \textcolor{red}{104} & \textcolor{red}{\textit{\texttt{cradle}}} & \textcolor{red}{0.081} & \textcolor{red}{\textit{\texttt{bus}}} & \textcolor{red}{0.112}\\ 
105 & \texttt{sea} & 0.106 & \texttt{water} & 0.058 & 106 & \texttt{rock} & 0.048 & \texttt{rock} & 0.06\\ 
110 & \texttt{painting} & 0.119 & \texttt{painting} & 0.06 & \textcolor{red}{112} & \textcolor{red}{\textit{\texttt{field}}} & \textcolor{red}{0.05} & \textcolor{red}{\textit{\texttt{bus}}} & \textcolor{red}{0.051}\\ 
113 & \texttt{table} & 0.116 & \texttt{table} & 0.066 & 115 & \texttt{plane} & 0.046 & \texttt{airplane} & 0.147\\ 
\textcolor{red}{120} & \textcolor{red}{\textit{\texttt{sidewalk}}} & \textcolor{red}{0.042} & \textcolor{red}{\textit{\texttt{track}}} & \textcolor{red}{0.075} & \textcolor{red}{125} & \textcolor{red}{\textit{\texttt{table}}} & \textcolor{red}{0.049} & \textcolor{red}{\textit{\texttt{wineglass}}} & \textcolor{red}{0.047}\\ 
\textcolor{red}{126} & \textcolor{red}{\textit{\texttt{stove}}} & \textcolor{red}{0.064} & \textcolor{red}{\textit{\texttt{bottle}}} & \textcolor{red}{0.163} & 127 & \texttt{book} & 0.104 & \texttt{book} & 0.096\\ 
\textcolor{red}{131} & \textcolor{red}{\textit{\texttt{signboard}}} & \textcolor{red}{0.043} & \textcolor{red}{\textit{\texttt{body}}} & \textcolor{red}{0.069} & \textcolor{red}{134} & \textcolor{red}{\textit{\texttt{bathtub}}} & \textcolor{red}{0.088} & \textcolor{red}{\textit{\texttt{boat}}} & \textcolor{red}{0.059}\\ 
\textcolor{red}{141} & \textcolor{red}{\textit{\texttt{skyscraper}}} & \textcolor{red}{0.065} & \textcolor{red}{\textit{\texttt{cage}}} & \textcolor{red}{0.068} & \textcolor{red}{155} & \textcolor{red}{\textit{\texttt{mountain}}} & \textcolor{red}{0.091} & \textcolor{red}{\textit{\texttt{train}}} & \textcolor{red}{0.058}\\ 
158 & \texttt{book} & 0.042 & \texttt{book} & 0.052 & 165 & \texttt{sea} & 0.051 & \texttt{water} & 0.051\\ 
168 & \texttt{railroad train} & 0.055 & \texttt{train} & 0.193 & \textcolor{red}{172} & \textcolor{red}{\textit{\texttt{car}}} & \textcolor{red}{0.055} & \textcolor{red}{\textit{\texttt{bus}}} & \textcolor{red}{0.101}\\ 
\textcolor{red}{173} & \textcolor{red}{\textit{\texttt{car}}} & \textcolor{red}{0.052} & \textcolor{red}{\textit{\texttt{bus}}} & \textcolor{red}{0.099} & 181 & \texttt{plant} & 0.068 & \texttt{pottedplant} & 0.14\\ 
\textcolor{red}{183} & \textcolor{red}{\textit{\texttt{person}}} & \textcolor{red}{0.041} & \textcolor{red}{\textit{\texttt{horse}}} & \textcolor{red}{0.187} & \textcolor{red}{184} & \textcolor{red}{\textit{\texttt{cradle}}} & \textcolor{red}{0.046} & \textcolor{red}{\textit{\texttt{cat}}} & \textcolor{red}{0.042}\\ 
\textcolor{red}{185} & \textcolor{red}{\textit{\texttt{chair}}} & \textcolor{red}{0.077} & \textcolor{red}{\textit{\texttt{horse}}} & \textcolor{red}{0.153} & \textcolor{red}{186} & \textcolor{red}{\textit{\texttt{person}}} & \textcolor{red}{0.051} & \textcolor{red}{\textit{\texttt{bird}}} & \textcolor{red}{0.094}\\ 
\textcolor{red}{191} & \textcolor{red}{\textit{\texttt{swimming pool}}} & \textcolor{red}{0.044} & \textcolor{red}{\textit{\texttt{pottedplant}}} & \textcolor{red}{0.072} & \textcolor{red}{198} & \textcolor{red}{\textit{\texttt{pool table}}} & \textcolor{red}{0.064} & \textcolor{red}{\textit{\texttt{ceiling}}} & \textcolor{red}{0.066}\\ 
\textcolor{red}{208} & \textcolor{red}{\textit{\texttt{shelf}}} & \textcolor{red}{0.047} & \textcolor{red}{\textit{\texttt{bus}}} & \textcolor{red}{0.062} & 211 & \texttt{computer} & 0.076 & \texttt{tvmonitor} & 0.089\\ 
\textcolor{red}{217} & \textcolor{red}{\textit{\texttt{toilet}}} & \textcolor{red}{0.049} & \textcolor{red}{\textit{\texttt{hair}}} & \textcolor{red}{0.055} & \textcolor{red}{218} & \textcolor{red}{\textit{\texttt{case}}} & \textcolor{red}{0.044} & \textcolor{red}{\textit{\texttt{track}}} & \textcolor{red}{0.165}\\ 
219 & \texttt{plane} & 0.065 & \texttt{airplane} & 0.189 & 220 & \texttt{road} & 0.066 & \texttt{road} & 0.066\\ 
222 & \texttt{grass} & 0.105 & \texttt{grass} & 0.046 & \textcolor{red}{223} & \textcolor{red}{\textit{\texttt{house}}} & \textcolor{red}{0.069} & \textcolor{red}{\textit{\texttt{airplane}}} & \textcolor{red}{0.055}\\ 
\textcolor{red}{231} & \textcolor{red}{\textit{\texttt{grandstand}}} & \textcolor{red}{0.097} & \textcolor{red}{\textit{\texttt{screen}}} & \textcolor{red}{0.047} & \textcolor{red}{234} & \textcolor{red}{\textit{\texttt{bridge}}} & \textcolor{red}{0.05} & \textcolor{red}{\textit{\texttt{train}}} & \textcolor{red}{0.042}\\ 
\textcolor{red}{239} & \textcolor{red}{\textit{\texttt{pool table}}} & \textcolor{red}{0.069} & \textcolor{red}{\textit{\texttt{horse}}} & \textcolor{red}{0.171} & 245 & \texttt{water} & 0.063 & \texttt{water} & 0.042\\ 
247 & \texttt{plane} & 0.079 & \texttt{airplane} & 0.177 & \textcolor{red}{248} & \textcolor{red}{\textit{\texttt{bed}}} & \textcolor{red}{0.127} & \textcolor{red}{\textit{\texttt{tvmonitor}}} & \textcolor{red}{0.063}\\ 
\textcolor{red}{251} & \textcolor{red}{\textit{\texttt{sofa}}} & \textcolor{red}{0.073} & \textcolor{red}{\textit{\texttt{pottedplant}}} & \textcolor{red}{0.053} & \textcolor{red}{257} & \textcolor{red}{\textit{\texttt{tent}}} & \textcolor{red}{0.042} & \textcolor{red}{\textit{\texttt{bus}}} & \textcolor{red}{0.279}\\ 
\textcolor{red}{260} & \textcolor{red}{\textit{\texttt{flower}}} & \textcolor{red}{0.082} & \textcolor{red}{\textit{\texttt{food}}} & \textcolor{red}{0.069} & \textcolor{red}{267} & \textcolor{red}{\textit{\texttt{apparel}}} & \textcolor{red}{0.042} & \textcolor{red}{\textit{\texttt{car}}} & \textcolor{red}{0.045}\\ 
276 & \texttt{earth} & 0.041 & \texttt{rock} & 0.047 & \textcolor{red}{278} & \textcolor{red}{\textit{\texttt{field}}} & \textcolor{red}{0.06} & \textcolor{red}{\textit{\texttt{sheep}}} & \textcolor{red}{0.044}\\ 
280 & \texttt{mountain} & 0.045 & \texttt{mountain} & 0.056 & 287 & \texttt{plant} & 0.078 & \texttt{pottedplant} & 0.07\\ 
\textcolor{red}{289} & \textcolor{red}{\textit{\texttt{pool table}}} & \textcolor{red}{0.049} & \textcolor{red}{\textit{\texttt{food}}} & \textcolor{red}{0.059} & 290 & \texttt{mountain} & 0.085 & \texttt{mountain} & 0.097\\ 
\textcolor{red}{293} & \textcolor{red}{\textit{\texttt{shelf}}} & \textcolor{red}{0.074} & \textcolor{red}{\textit{\texttt{bottle}}} & \textcolor{red}{0.105} & \textcolor{red}{298} & \textcolor{red}{\textit{\texttt{path}}} & \textcolor{red}{0.047} & \textcolor{red}{\textit{\texttt{motorbike}}} & \textcolor{red}{0.068}\\ 
\textcolor{red}{305} & \textcolor{red}{\textit{\texttt{waterfall}}} & \textcolor{red}{0.057} & \textcolor{red}{\textit{\texttt{mountain}}} & \textcolor{red}{0.047} & \textcolor{red}{309} & \textcolor{red}{\textit{\texttt{washer}}} & \textcolor{red}{0.109} & \textcolor{red}{\textit{\texttt{bus}}} & \textcolor{red}{0.065}\\ 
318 & \texttt{computer} & 0.079 & \texttt{tvmonitor} & 0.251 & \textcolor{red}{322} & \textcolor{red}{\textit{\texttt{ball}}} & \textcolor{red}{0.054} & \textcolor{red}{\textit{\texttt{sheep}}} & \textcolor{red}{0.044}\\ 
\textcolor{red}{324} & \textcolor{red}{\textit{\texttt{mountain}}} & \textcolor{red}{0.071} & \textcolor{red}{\textit{\texttt{motorbike}}} & \textcolor{red}{0.048} & 325 & \texttt{person} & 0.04 & \texttt{head} & 0.059\\ 
\textcolor{red}{327} & \textcolor{red}{\textit{\texttt{waterfall}}} & \textcolor{red}{0.055} & \textcolor{red}{\textit{\texttt{bird}}} & \textcolor{red}{0.087} & \textcolor{red}{337} & \textcolor{red}{\textit{\texttt{water}}} & \textcolor{red}{0.072} & \textcolor{red}{\textit{\texttt{boat}}} & \textcolor{red}{0.109}\\ 
\textcolor{red}{341} & \textcolor{red}{\textit{\texttt{sea}}} & \textcolor{red}{0.153} & \textcolor{red}{\textit{\texttt{boat}}} & \textcolor{red}{0.076} & 344 & \texttt{person} & 0.052 & \texttt{person} & 0.048\\ 
345 & \texttt{autobus} & 0.042 & \texttt{bus} & 0.142 & \textcolor{red}{347} & \textcolor{red}{\textit{\texttt{palm}}} & \textcolor{red}{0.051} & \textcolor{red}{\textit{\texttt{bicycle}}} & \textcolor{red}{0.083}\\ 
348 & \texttt{mountain} & 0.058 & \texttt{mountain} & 0.125 & \textcolor{red}{354} & \textcolor{red}{\textit{\texttt{cradle}}} & \textcolor{red}{0.042} & \textcolor{red}{\textit{\texttt{chair}}} & \textcolor{red}{0.053}\\ 
\textcolor{red}{357} & \textcolor{red}{\textit{\texttt{rock}}} & \textcolor{red}{0.058} & \textcolor{red}{\textit{\texttt{sheep}}} & \textcolor{red}{0.061} & \textcolor{red}{360} & \textcolor{red}{\textit{\texttt{pool table}}} & \textcolor{red}{0.048} & \textcolor{red}{\textit{\texttt{bird}}} & \textcolor{red}{0.041}\\ 
364 & \texttt{field} & 0.058 & \texttt{plant} & 0.041 & 372 & \texttt{work surface} & 0.045 & \texttt{cabinet} & 0.049\\ 
\textcolor{red}{379} & \textcolor{red}{\textit{\texttt{bridge}}} & \textcolor{red}{0.092} & \textcolor{red}{\textit{\texttt{bus}}} & \textcolor{red}{0.046} & \textcolor{red}{383} & \textcolor{red}{\textit{\texttt{bed}}} & \textcolor{red}{0.069} & \textcolor{red}{\textit{\texttt{curtain}}} & \textcolor{red}{0.079}\\ 
\textcolor{red}{384} & \textcolor{red}{\textit{\texttt{washer}}} & \textcolor{red}{0.043} & \textcolor{red}{\textit{\texttt{bicycle}}} & \textcolor{red}{0.201} & 386 & \texttt{autobus} & 0.067 & \texttt{bus} & 0.200\\ 
\textcolor{red}{387} & \textcolor{red}{\textit{\texttt{hovel}}} & \textcolor{red}{0.04} & \textcolor{red}{\textit{\texttt{train}}} & \textcolor{red}{0.085} & 389 & \texttt{chair} & 0.066 & \texttt{chair} & 0.051\\ 
398 & \texttt{windowpane} & 0.073 & \texttt{windowpane} & 0.07 & 400 & \texttt{plant} & 0.043 & \texttt{pottedplant} & 0.097\\ 
\textcolor{red}{408} & \textcolor{red}{\textit{\texttt{toilet}}} & \textcolor{red}{0.045} & \textcolor{red}{\textit{\texttt{bottle}}} & \textcolor{red}{0.099} & \textcolor{red}{412} & \textcolor{red}{\textit{\texttt{bed}}} & \textcolor{red}{0.079} & \textcolor{red}{\textit{\texttt{airplane}}} & \textcolor{red}{0.086}\\ 
\textcolor{red}{413} & \textcolor{red}{\textit{\texttt{pool table}}} & \textcolor{red}{0.09} & \textcolor{red}{\textit{\texttt{motorbike}}} & \textcolor{red}{0.07} & \textcolor{red}{415} & \textcolor{red}{\textit{\texttt{seat}}} & \textcolor{red}{0.044} & \textcolor{red}{\textit{\texttt{tvmonitor}}} & \textcolor{red}{0.045}\\ 
417 & \texttt{sand} & 0.06 & \texttt{sand} & 0.049 & \textcolor{red}{419} & \textcolor{red}{\textit{\texttt{bed}}} & \textcolor{red}{0.061} & \textcolor{red}{\textit{\texttt{tvmonitor}}} & \textcolor{red}{0.054}\\ 
\textcolor{red}{422} & \textcolor{red}{\textit{\texttt{seat}}} & \textcolor{red}{0.089} & \textcolor{red}{\textit{\texttt{tvmonitor}}} & \textcolor{red}{0.056} & 430 & \texttt{bed} & 0.078 & \texttt{bedclothes} & 0.042\\ 
\textcolor{red}{434} & \textcolor{red}{\textit{\texttt{case}}} & \textcolor{red}{0.047} & \textcolor{red}{\textit{\texttt{cup}}} & \textcolor{red}{0.041} & 435 & \texttt{runway} & 0.072 & \texttt{airplane} & 0.189\\ 
438 & \texttt{plane} & 0.045 & \texttt{airplane} & 0.235 & \textcolor{red}{444} & \textcolor{red}{\textit{\texttt{sofa}}} & \textcolor{red}{0.045} & \textcolor{red}{\textit{\texttt{plant}}} & \textcolor{red}{0.09}\\ 
445 & \texttt{car} & 0.201 & \texttt{car} & 0.093 & \textcolor{red}{446} & \textcolor{red}{\textit{\texttt{pool table}}} & \textcolor{red}{0.193} & \textcolor{red}{\textit{\texttt{tvmonitor}}} & \textcolor{red}{0.086}\\ 
454 & \texttt{car} & 0.218 & \texttt{car} & 0.156 & 463 & \texttt{snow} & 0.059 & \texttt{snow} & 0.118\\ 
465 & \texttt{crosswalk} & 0.097 & \texttt{road} & 0.047 & \textcolor{red}{475} & \textcolor{red}{\textit{\texttt{cradle}}} & \textcolor{red}{0.061} & \textcolor{red}{\textit{\texttt{train}}} & \textcolor{red}{0.132}\\ 
\textcolor{red}{477} & \textcolor{red}{\textit{\texttt{desk}}} & \textcolor{red}{0.104} & \textcolor{red}{\textit{\texttt{tvmonitor}}} & \textcolor{red}{0.085} & 480 & \texttt{sofa} & 0.086 & \texttt{sofa} & 0.081\\ 
\textcolor{red}{483} & \textcolor{red}{\textit{\texttt{swivel chair}}} & \textcolor{red}{0.052} & \textcolor{red}{\textit{\texttt{horse}}} & \textcolor{red}{0.041} & 484 & \texttt{water} & 0.15 & \texttt{water} & 0.102\\ 
\textcolor{red}{485} & \textcolor{red}{\textit{\texttt{sofa}}} & \textcolor{red}{0.056} & \textcolor{red}{\textit{\texttt{airplane}}} & \textcolor{red}{0.045} & 500 & \texttt{sofa} & 0.156 & \texttt{sofa} & 0.11\\ 
\textcolor{red}{502} & \textcolor{red}{\textit{\texttt{washer}}} & \textcolor{red}{0.07} & \textcolor{red}{\textit{\texttt{train}}} & \textcolor{red}{0.134} & 503 & \texttt{bookcase} & 0.109 & \texttt{book} & 0.075\\ 
509 & \texttt{computer} & 0.044 & \texttt{tvmonitor} & 0.074 & \\
\bottomrule
\end{tabular}
    }
    \caption{We show labels for all neurons from the penultimate layer of a ResNet18 model that are marked as highly activated by both datasets by NetDissect~\cite{bau2017netdissect}. We find that a 69/123 of neurons correspond to labels that are radically different (shown in \textcolor{red}{\textit{\texttt{red}}}). The remainder correspond to either the same or very similar concepts.}
    \label{tab:netdissect_supp}
\end{table*}

As mentioned by Fong \etal~\cite{fong2018net2vec} and Olah \etal~\cite{olah2020zoom}, neurons in deep neural networks can be \emph{poly-semantic}, i.e, some neurons can recognize multiple concepts. We check if the results from above are due to such neurons, and confirm that that is not the case: out of the 69 neurons, only 7 are highly activated (IOU>0.04) by both concepts. Table~\ref{tab:polysemantic_supp} contains the IOU scores for both the ADE20k and Pascal label for each neuron outputting very different concepts. 

\begin{table*}[ht]
    \centering
    
    \resizebox{0.8\textwidth}{!}{%
    \begin{tabular}{ccccccc}
 \toprule
\multicolumn{1}{l}{} &  &  & \multicolumn{2}{c}{Probe dataset: ADE20k} & \multicolumn{2}{c}{Probe dataset: Pascal} \\
\cmidrule(l{4pt}r{4pt}){4-5} \cmidrule(l{4pt}r{4pt}){6-7}
\multicolumn{1}{l}{\multirow{-2}{*}{neuron}} & \multirow{-2}{*}{ADE20k label} & \multirow{-2}{*}{Pascal label} & \multicolumn{1}{l}{IOU ADE20k label} & \multicolumn{1}{l}{IOU Pascal label} & \multicolumn{1}{l}{IOU ADE20k label} & \multicolumn{1}{l}{IOU Pascal label} \\
\midrule
1 & \texttt{counter} & \texttt{bottle} & 0.059 & 0.006 & 0.006 & 0.049\\ 
4 & \texttt{seat} & \texttt{tvmonitor} & 0.064 & 0.0 & 0.0 & 0.074\\ 
22 & \texttt{bookcase} & \texttt{bus} & 0.07 & 0.0 & 0.0 & 0.048\\ 
47 & \texttt{pool table} & \texttt{airplane} & 0.135 & 0.0 & 0.002 & 0.079\\ 
63 & \texttt{field} & \texttt{muzzle} & 0.053 & 0.0 & 0.0 & 0.042\\ 
\textcolor{red}{\textit{73}} & \textcolor{red}{\textit{\texttt{water}}} & \textcolor{red}{\textit{\texttt{bird}}} & \textcolor{red}{\textit{0.041}} & \textcolor{red}{\textit{0.002}} & \textcolor{red}{\textit{0.052}} & \textcolor{red}{\textit{0.08}}\\ 
102 & \texttt{bathtub} & \texttt{cat} & 0.04 & 0.0 & 0.0 & 0.055\\ 
104 & \texttt{cradle} & \texttt{bus} & 0.081 & 0.0 & 0.0 & 0.112\\ 
112 & \texttt{field} & \texttt{bus} & 0.05 & 0.0 & 0.0 & 0.051\\ 
120 & \texttt{sidewalk} & \texttt{track} & 0.042 & 0.001 & 0.023 & 0.075\\ 
\textcolor{red}{\textit{125}} & \textcolor{red}{\textit{\texttt{table}}} & \textcolor{red}{\textit{\texttt{wineglass}}} & \textcolor{red}{\textit{0.049}} & \textcolor{red}{\textit{0.0}} & \textcolor{red}{\textit{0.043}} & \textcolor{red}{\textit{0.047}} \\ 
126 & \texttt{stove} & \texttt{bottle} & 0.064 & 0.029 & 0.005 & 0.163\\ 
\textcolor{red}{\textit{131}} & \textcolor{red}{\textit{\texttt{signboard}}} & \textcolor{red}{\textit{\texttt{body}}} & \textcolor{red}{\textit{0.043}} & \textcolor{red}{\textit{0.0}} & \textcolor{red}{\textit{0.06}} & \textcolor{red}{\textit{0.069}}\\ 
134 & \texttt{bathtub} & \texttt{boat} & 0.088 & 0.001 & 0.005 & 0.059\\ 
141 & \texttt{skyscraper} & \texttt{cage} & 0.065 & 0.001 & 0.0 & 0.068\\ 
155 & \texttt{mountain} & \texttt{train} & 0.091 & 0.0 & 0.038 & 0.058\\ 
172 & \texttt{car} & \texttt{bus} & 0.055 & 0.0 & 0.015 & 0.101\\ 
173 & \texttt{car} & \texttt{bus} & 0.052 & 0.0 & 0.013 & 0.099\\ 
183 & \texttt{person} & \texttt{horse} & 0.041 & 0.016 & 0.003 & 0.187\\ 
184 & \texttt{cradle} & \texttt{cat} & 0.046 & 0.0 & 0.0 & 0.042\\ 
185 & \texttt{chair} & \texttt{horse} & 0.077 & 0.014 & 0.011 & 0.153\\ 
186 & \texttt{person} & \texttt{bird} & 0.051 & 0.001 & 0.017 & 0.094\\ 
191 & \texttt{swimming pool} & \texttt{pottedplant} & 0.044 & 0.0 & 0.0 & 0.072\\ 
198 & \texttt{pool table} & \texttt{ceiling} & 0.064 & 0.035 & 0.001 & 0.066\\ 
208 & \texttt{shelf} & \texttt{bus} & 0.047 & 0.0 & 0.0 & 0.062\\ 
217 & \texttt{toilet} & \texttt{hair} & 0.049 & 0.001 & 0.0 & 0.055\\ 
218 & \texttt{case} & \texttt{track} & 0.044 & 0.001 & 0.0 & 0.165\\ 
223 & \texttt{house} & \texttt{airplane} & 0.069 & 0.0 & 0.0 & 0.055\\ 
231 & \texttt{grandstand} & \texttt{screen} & 0.097 & 0.0 & 0.007 & 0.047\\ 
234 & \texttt{bridge} & \texttt{train} & 0.05 & 0.0 & 0.014 & 0.042\\ 
239 & \texttt{pool table} & \texttt{horse} & 0.069 & 0.011 & 0.0 & 0.171\\ 
248 & \texttt{bed} & \texttt{tvmonitor} & 0.127 & 0.0 & 0.027 & 0.063\\ 
251 & \texttt{sofa} & \texttt{pottedplant} & 0.073 & 0.0 & 0.033 & 0.053\\ 
257 & \texttt{tent} & \texttt{bus} & 0.042 & 0.0 & 0.005 & 0.279\\ 
\textcolor{red}{\textit{260}} & \textcolor{red}{\textit{\texttt{flower}}} & \textcolor{red}{\textit{\texttt{food}}} & \textcolor{red}{\textit{0.082}} & \textcolor{red}{\textit{0.033}} & \textcolor{red}{\textit{0.064}} & \textcolor{red}{\textit{0.069}}\\ 
267 & \texttt{apparel} & \texttt{car} & 0.042 & 0.023 & 0.0 & 0.045\\ 
278 & \texttt{field} & \texttt{sheep} & 0.06 & 0.0 & 0.0 & 0.044\\ 
289 & \texttt{pool table} & \texttt{food} & 0.049 & 0.024 & 0.0 & 0.059\\ 
293 & \texttt{shelf} & \texttt{bottle} & 0.074 & 0.025 & 0.0 & 0.105\\ 
298 & \texttt{path} & \texttt{motorbike} & 0.047 & 0.0 & 0.0 & 0.068\\ 
\textcolor{red}{\textit{305}} & \textcolor{red}{\textit{\texttt{waterfall}}} & \textcolor{red}{\textit{\texttt{mountain}}} & \textcolor{red}{\textit{0.057}} & \textcolor{red}{\textit{0.049}} & \textcolor{red}{\textit{0.0}} & \textcolor{red}{\textit{0.047}}\\ 
309 & \texttt{washer} & \texttt{bus} & 0.109 & 0.0 & 0.013 & 0.065\\ 
322 & \texttt{ball} & \texttt{sheep} & 0.054 & 0.0 & 0.005 & 0.044\\ 
324 & \texttt{mountain} & \texttt{motorbike} & 0.071 & 0.0 & 0.015 & 0.048\\ 
327 & \texttt{waterfall} & \texttt{bird} & 0.055 & 0.001 & 0.0 & 0.087\\ 
\textcolor{red}{\textit{337}} & \textcolor{red}{\textit{\texttt{water}}} & \textcolor{red}{\textit{\texttt{boat}}} & \textcolor{red}{\textit{0.072}} & \textcolor{red}{\textit{0.031}} & \textcolor{red}{\textit{0.053}} & \textcolor{red}{\textit{0.109}}\\ 
341 & \texttt{sea} & \texttt{boat} & 0.153 & 0.014 & 0.0 & 0.076\\ 
347 & \texttt{palm} & \texttt{bicycle} & 0.051 & 0.001 & 0.0 & 0.083\\ 
354 & \texttt{cradle} & \texttt{chair} & 0.042 & 0.03 & 0.0 & 0.053\\ 
357 & \texttt{rock} & \texttt{sheep} & 0.058 & 0.0 & 0.006 & 0.061\\ 
360 & \texttt{pool table} & \texttt{bird} & 0.048 & 0.0 & 0.0 & 0.041\\ 
379 & \texttt{bridge} & \texttt{bus} & 0.092 & 0.0 & 0.03 & 0.046\\ 
\textcolor{red}{\textit{383}} & \textcolor{red}{\textit{\texttt{bed}}} & \textcolor{red}{\textit{\texttt{curtain}}} & \textcolor{red}{\textit{0.069}} & \textcolor{red}{\textit{0.064}} & \textcolor{red}{\textit{0.01}} & \textcolor{red}{\textit{0.079}}\\ 
384 & \texttt{washer} & \texttt{bicycle} & 0.043 & 0.018 & 0.0 & 0.201\\ 
387 & \texttt{hovel} & \texttt{train} & 0.04 & 0.0 & 0.0 & 0.085\\ 
408 & \texttt{toilet} & \texttt{bottle} & 0.045 & 0.002 & 0.0 & 0.099\\ 
412 & \texttt{bed} & \texttt{airplane} & 0.079 & 0.0 & 0.008 & 0.086\\ 
413 & \texttt{pool table} & \texttt{motorbike} & 0.09 & 0.0 & 0.003 & 0.07\\ 
415 & \texttt{seat} & \texttt{tvmonitor} & 0.044 & 0.0 & 0.0 & 0.045\\ 
419 & \texttt{bed} & \texttt{tvmonitor} & 0.061 & 0.0 & 0.016 & 0.054\\ 
422 & \texttt{seat} & \texttt{tvmonitor} & 0.089 & 0.0 & 0.0 & 0.056\\ 
434 & \texttt{case} & \texttt{cup} & 0.047 & 0.001 & 0.0 & 0.041\\ 
444 & \texttt{sofa} & \texttt{plant} & 0.045 & 0.009 & 0.014 & 0.09\\ 
446 & \texttt{pool table} & \texttt{tvmonitor} & 0.193 & 0.0 & 0.006 & 0.086\\ 
475 & \texttt{cradle} & \texttt{train} & 0.061 & 0.0 & 0.0 & 0.132\\ 
477 & \texttt{desk} & \texttt{tvmonitor} & 0.104 & 0.0 & 0.0 & 0.085\\ 
483 & \texttt{swivel chair} & \texttt{horse} & 0.052 & 0.006 & 0.0 & 0.041\\ 
485 & \texttt{sofa} & \texttt{airplane} & 0.056 & 0.0 & 0.024 & 0.045\\ 
502 & \texttt{washer} & \texttt{train} & 0.07 & 0.0 & 0.006 & 0.134\\ 
\bottomrule
\end{tabular}

    }
    \caption{For all neurons from Tab.~\ref{tab:netdissect_supp} that output radically different concepts when explanations are computed using ADE20k vs Pascal, we compute the IOU scores for the other concept as well. Other than the 7 attributes marked in \textcolor{red}{\texttt{\textit{red}}}, the IOU scores are all below 0.04, suggesting that this is not because the neurons are polysemantic. }
    \label{tab:polysemantic_supp}
\end{table*}

\smallsec{TCAV}
We report the cosine similarities between the concept activation vectors learned using ADE20k and Pascal as probe datasets for all 32 concepts that have a base rate of at least 1\% in Table~\ref{tab:tcav_supp}. On the whole, we see that the vectors are not very similar, despite the vectors predicting the concepts well. 
\begin{table*}[t]
    \centering
    
    \resizebox{\linewidth}{!}{%
\begin{tabular}{cccc|cccc}
\toprule
Concept & ADE20k AUC & Pascal AUC & Cos.sim. & Concept & ADE20k AUC & Pascal AUC & Cos.sim. \\
\midrule
\texttt{bag} & 79.4 & 75.4 & 0.006 & \texttt{book} & 90.4 & 84.6 & 0.138 \\ 
 \texttt{bottle} & 88.5 & 85.6 & 0.035 & \texttt{box} & 83.0 & 80.1 & 0.086 \\ 
 \texttt{building} & 97.4 & 90.0 & 0.161 & \texttt{cabinet} & 91.3 & 92.4 & 0.03 \\ 
 \texttt{car} & 96.9 & 90.3 & 0.147 & \texttt{ceiling} & 96.6 & 93.0 & 0.267 \\ 
 \texttt{chair} & 90.5 & 89.6 & 0.034 & \texttt{curtain} & 91.6 & 89.5 & 0.112 \\ 
 \texttt{door} & 81.5 & 87.8 & 0.134 & \texttt{fence} & 86.1 & 84.7 & 0.09 \\ 
 \texttt{floor} & 97.4 & 92.1 & 0.208 & \texttt{grass} & 95.1 & 91.7 & 0.04 \\ 
 \texttt{light} & 92.4 & 85.0 & 0.043 & \texttt{mountain} & 94.2 & 90.8 & 0.02 \\ 
 \texttt{painting} & 94.8 & 91.4 & 0.116 & \texttt{person} & 92.2 & 92.1 & 0.253 \\ 
 \texttt{plate} & 90.6 & 94.8 & -0.009 & \texttt{pole} & 89.0 & 79.3 & 0.059 \\ 
 \texttt{pot} & 79.3 & 85.2 & 0.142 & \texttt{road} & 98.0 & 91.8 & 0.041 \\ 
 \texttt{rock} & 92.6 & 82.8 & -0.024 & \texttt{sidewalk} & 97.0 & 92.5 & 0.071 \\ 
 \texttt{signboard} & 90.6 & 76.5 & 0.091 & \texttt{sky} & 98.9 & 79.8 & 0.104 \\ 
 \texttt{sofa} & 95.9 & 91.2 & -0.009 & \texttt{table} & 93.4 & 93.5 & 0.06 \\ 
 \texttt{tree} & 96.8 & 89.2 & 0.172 & \texttt{wall} & 95.9 & 91.3 & 0.027 \\ 
 \texttt{water} & 95.2 & 94.6 & 0.078 & \texttt{windowpane} & 91.5 & 90.1 & 0.078 \\ 
 \bottomrule
\end{tabular}}
\caption{We report the cosine similarities between the concept activation vectors learned using ADE20k and Pascal datasets. In general, the vectors learned from different datasets do not correlate well. }
    \label{tab:tcav_supp}
\end{table*}

\subsection{Difference in probe dataset distribution} 
\label{sec:data_dist_supp}
\begin{figure*}[t]
    \centering
    \includegraphics[width=\textwidth]{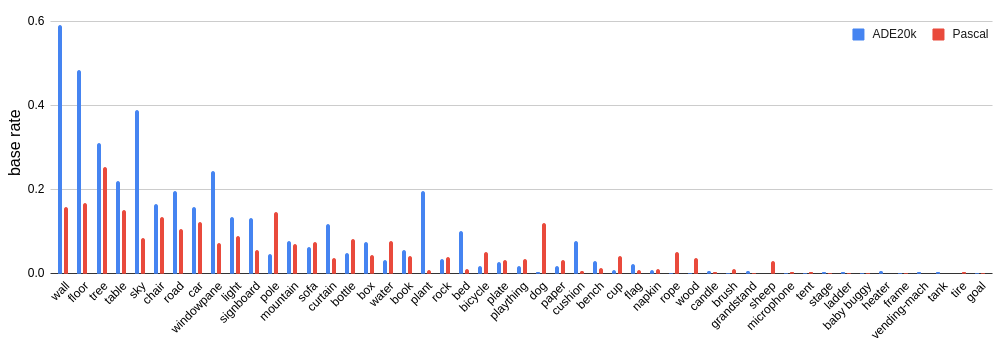}
    \caption{Different concepts have very different base rates across Pascal and ADE20k. The graph shows the base rates for the different concepts highlighted within Table 2 in the main paper. }
    \label{fig:base_rates}
\end{figure*}

The first method we use to look at the difference in the 2 probe datasets we used was to consider the base rates of different concepts within the dataset. As noted in Section 3 of the main paper, there are some sizable differences. Figure~\ref{fig:base_rates} contains the base rates for all concepts highlighted in Table 2 of the main paper. Some concepts that have very different base rates are \texttt{wall} (highlighted for \texttt{bow-window} when using ADE20k, but not Pascal), \texttt{floor} (highlighted for \texttt{auto-showroom} when using ADE20k but not Pascal), \texttt{dog} (highlighted for \texttt{corn-field} when using Pascal, but not ADE20k) and \texttt{pole} (highlighted for \texttt{hardware-store} for Pascal, but not ADE20k). 

\begin{figure*}[t]
    \centering
    \includegraphics[width=\textwidth]{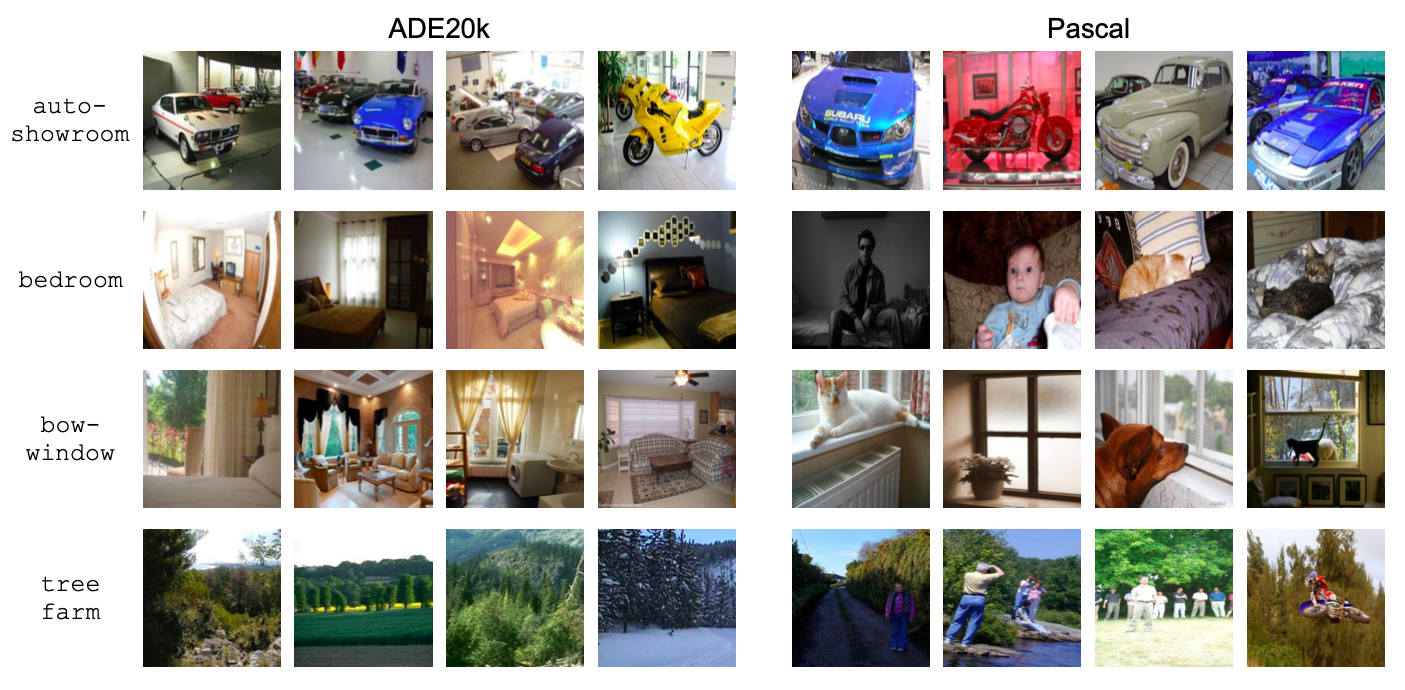}
    \caption{We view a few example images from ADE20k and Pascal for 4 scene classes that had very different explanations in Table 2 from the original paper. We see that these classes have very different distributions; for example, the images labelled as \texttt{bedroom} from the Pascal dataset tend to have an animal or person on a bed, whereas the ones from ADE20k do not.}
    \label{fig:dset_distribution_supp}
\end{figure*}

However, more than just the base rate, the images themselves look very different across scenes. We visualize random images from different scenes in Figure~\ref{fig:dset_distribution_supp}, and find, for example, images labelled \texttt{bedroom} in Pascal tend to have either a person or animal sleeping on a bed, without much of the remaining bedroom being shown, whereas ADE20k features images of full bedrooms. Similarly, images labelled \texttt{tree-farm} contain people in Pascal, but do not in ADE20k. 

\smallsec{Upper bounds}

Finally, we present a simple method to compare the similarity of the probe dataset with that of the training dataset by noting that the probe dataset establishes a strict \emph{upper bound} on the fraction of the model that can be explained. This is intuitively true since the set of semantic labeled concepts is finite, but actually goes deeper than that. Consider the following experiment: we take the original black-box model, run it on a probe dataset to make predictions, and then train a new classifier to emulate those predictions. If this classifier is restricted to use only the labeled concepts then this is similar to a concept-based explanation. However, even if it's trained on the rich underlying visual features it would not perform perfectly due to the differences between the original training dataset and the probe dataset. 

Concretely, consider a black-box ResNet18-based~\cite{he2016resnet}  model trained on the Places365~\cite{zhou2017places} dataset. We reset and re-train its final linear classification layer on the Pascal~\cite{Everingham10pascal}  probe dataset to emulate the original scene predictions; this achieves only 63.7\% accuracy. Similarly, on the ADE20k~\cite{zhou2017ade20k,zhou2019ade20k_ijcv} as the probe dataset it achieves only slightly better 75.7\% accuracy, suggesting that this dataset is somewhat more similar to Places365 than Pascal but still far from fully capturing the distribution. This is not to suggest that the only way to generate concept-based explanations is to collect concept labels for the original training set (which may lead to overfitting); rather, it's important to acknowledge this limitation and quantify the explanation method based on such upper bounds.

Similarly, we can ask how well the Concept Bottleneck model~\cite{koh2020conceptbottleneck} can be explained using the CUB test dataset. However, in this case, since the training and test distributions are (hopefully!) similar, we would expect our upper bound to be reasonably high. We check this with our same set up, and find that this is indeed the case -- resetting and retraining the final linear layer, using the model's predictions as our targets achieves an accuracy of 89.3\%.

\section{Concepts used: more details}
\label{sec:supp-corr}
\begin{figure*}[t]
\begin{center}
\begin{minipage}{0.45\textwidth}
    \centering
    \includegraphics[width=\textwidth]{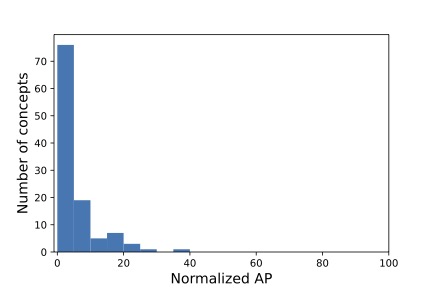}
\end{minipage}
\begin{minipage}{0.45\textwidth}
    \centering
    \includegraphics[width=\textwidth]{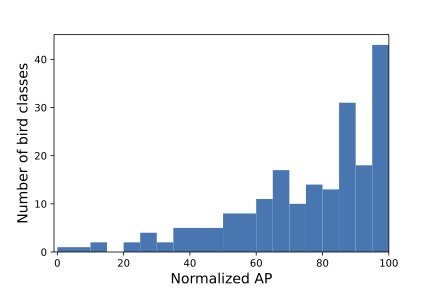}
\end{minipage}
    \caption{We compare the normalized APs when trying to learn CUB concepts (\emph{left}) to the normalized APs of the CUB target classes for the Concept Bottleneck model(\emph{right}). On average, the concepts are much harder to learn.}
    \label{fig:cub_learnability_supp}
\end{center}
\end{figure*}

Here, we provide additional results regarding learning CUB concepts from Section 4.2 of the main paper. The CUB dataset was used by Concept Bottleneck~\cite{koh2020conceptbottleneck}, an interpretable-by-design model. This method learned the concepts as an intermediate layer within the network, and then used these concepts to pretdict the target class. Figure~\ref{fig:cub_learnability_supp} contains the histograms of the normalized AP scores for the 112 concepts from CUB~\cite{WahCUB_200_2011} as well as the APs for the target bird classes learned by the model. Similar to learning classifiers for the Broden~\cite{bau2017netdissect} concepts, we learn a linear model using features from an Imagenet~\cite{russakovsky2015imagenet} trained Resnet18~\cite{he2016resnet} model. On average, we see that the bird classes are much better learned than the concepts.

\section{Human study details}
\label{sec:supp-humanstudy}

In Section 5 of the main paper, we discuss the human studies we ran to understand how well humans are able to reason about concept-based explanations as the number of concepts used within the explanation increases. 
In this section, we provide additional details.

To recap, we compare four types of explanations: (1) concept-based explanations that use 8 concepts, (2) concept-based explanations that use 16 concepts, (3) concept-based explanations that use 32 concepts, and (4) example-based explanations that consist of 10 example images for which the model predicts a certain class.
(4) is a baseline that doesn't use concepts.

For a fair comparison, all four types of explanations are evaluated on the same inputs.
We generate five sets of input where each set consists of 5 images from one scene group (commercial buildings, shops, markets, cities, and towns) and 5 images from another scene group (home or hotel). 
Recall that these are images where the model output match the explanation output (i.e., the class with the highest explanation score calculated based on ground-truth concept labels).
Hence, if the participants correctly identify all concepts that appear in a given image, they are guaranteed to get the highest explanation score for the model output class.

To reduce the variance with respect to the input, we had 5 participants for each set of input and explanation type.
For 32 concepts explanations, each participant saw 5 images from only one of the two scene groups because the study got too long and overwhelming with the full set of 10 images.
For all other explanations, each participant saw the full set of 10 images.
In total, we had 125 participants: 50 participants for the study with 32 concepts explanations and 25 participants for the other three studies.
Each participant sees only one type of explanation as we conduct a between-group study.

More specifically, we recruited participants through Amazon Mechanical Turk who are US-based, have done over 1000 Human Intelligence Tasks, and have prior approval rate of at least 98\%. 
The demographic distribution was: man 59\%, woman 41\%; no race/ethnicity reported 82\%, White 17\%, Black/African American 1\%, Asian 1\%.
The self-reported machine learning experience was 2.5 $\pm$ 1.0, between ``2: have heard about...'' and ``3: know the basics...'' 
We did not collect any personally identifiable information. 
Participants were compensated based on the state-level minimum wage of \$12/hr. In total, $\sim$\$800 was spent on running human studies.


\section{User interface snapshots}
\label{sec:supp-ui}

In Section 5.1 of the main paper, we outlined our human study design.\footnote{We note that much of our study design and UI is based on the recent work by Kim et al.~\cite{kim2021hive} who propose a human evaluation framework called HIVE for evaluating visual interpretability methods.}
Here we provide snapshots of our study UIs in the following order.

\begin{figure*}[t!]
\centering
\includegraphics[width=0.8\linewidth]{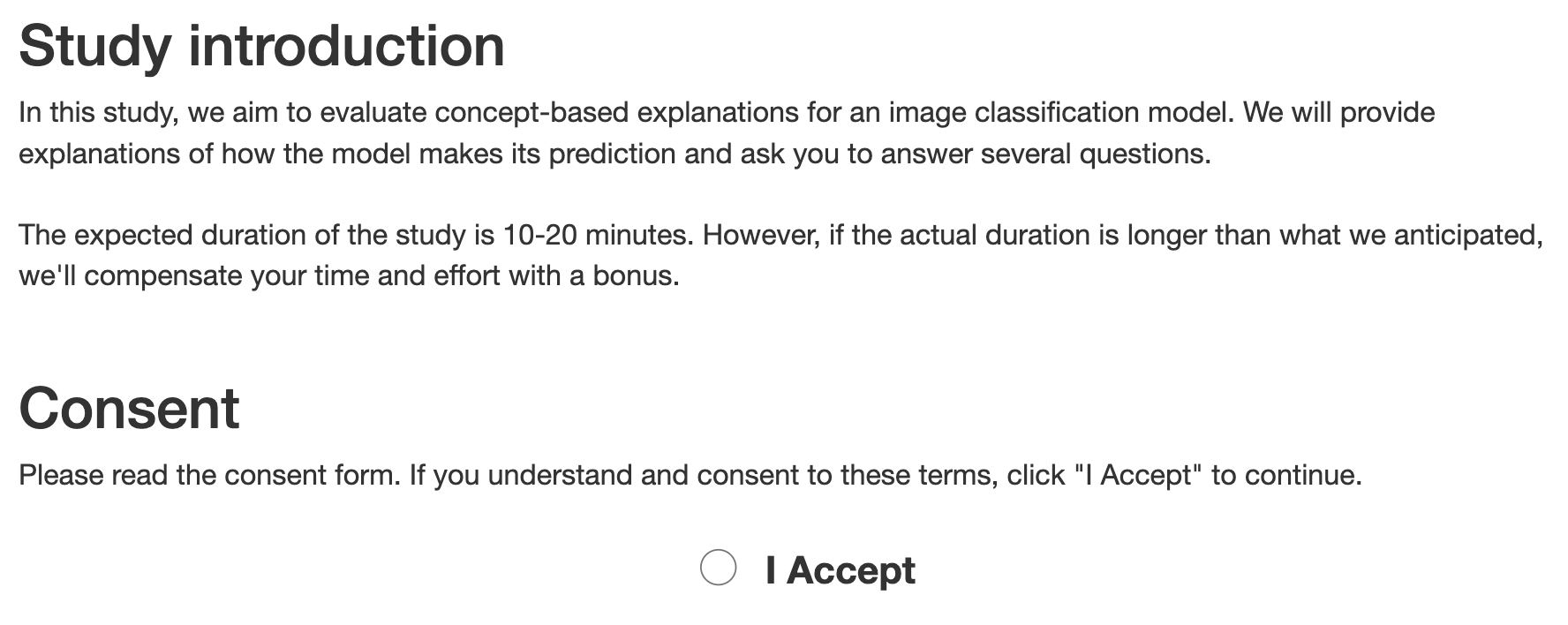}
\caption{\textbf{UI - Study introduction}}
\label{fig:ui_study_intro}
\vspace{2cm}
\includegraphics[width=0.75\linewidth]{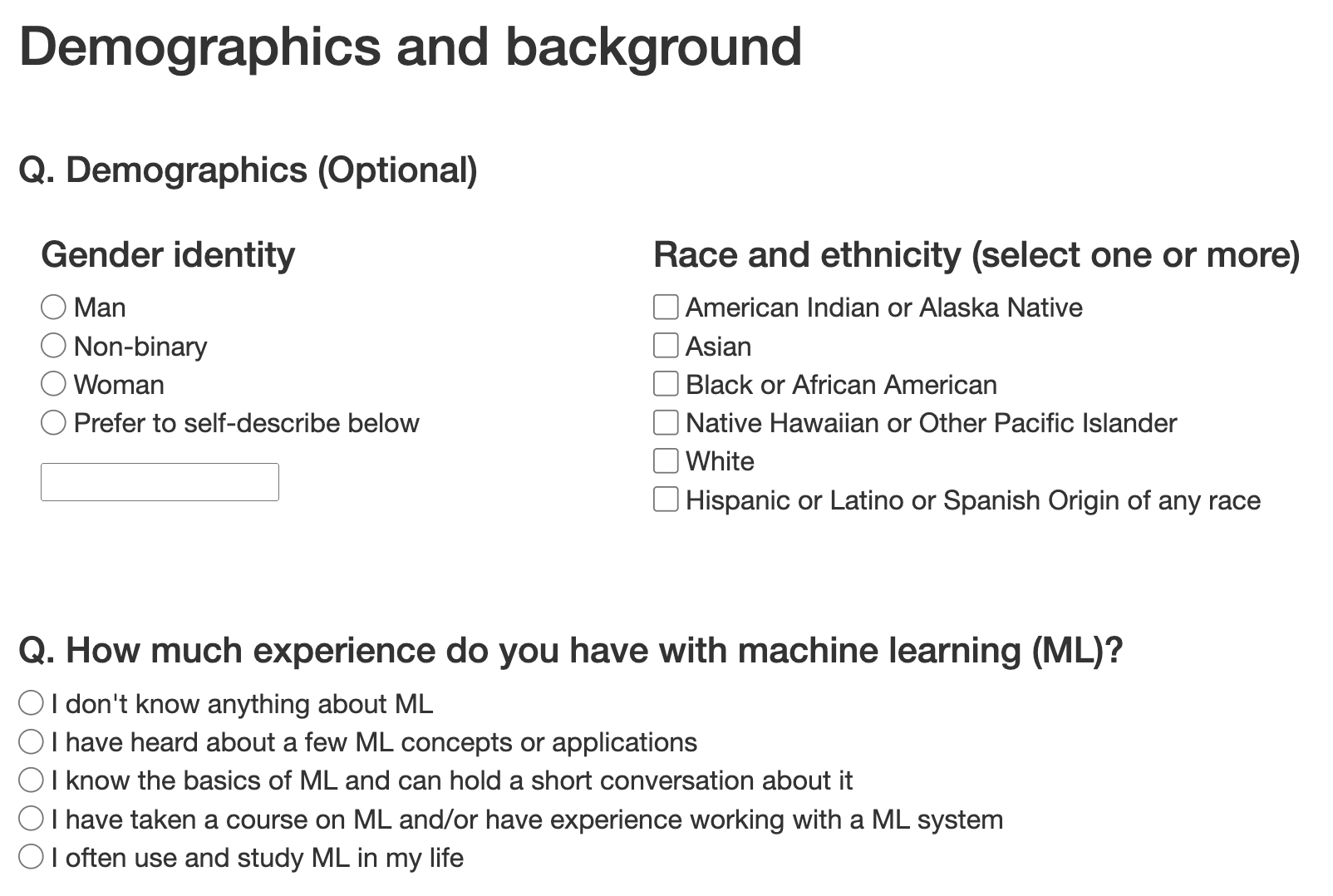}
\caption{\textbf{UI - Demographics and background}}
\label{fig:ui_demographic_background}
\end{figure*}

\smallsec{Study introduction.} For each participant, we introduce the study, present a consent form, and receive informed consent for participation in the study. The consent form was approved by our institution's Institutional Review Board and acknowledges that participation is voluntary, refusal to participate will involve no penalty or loss of benefits, etc.
See~\cref{fig:ui_study_intro}.

\smallsec{Demographics and background.} Following HIVE~\cite{kim2022hive}, we request optional demographic data regarding gender identity, race and ethnicity, as well as the participant's experience with machine learning. We collect this information to help future researchers calibrate our results.
See~\cref{fig:ui_demographic_background}.

\begin{figure*}[t!]
\centering
\includegraphics[width=\linewidth]{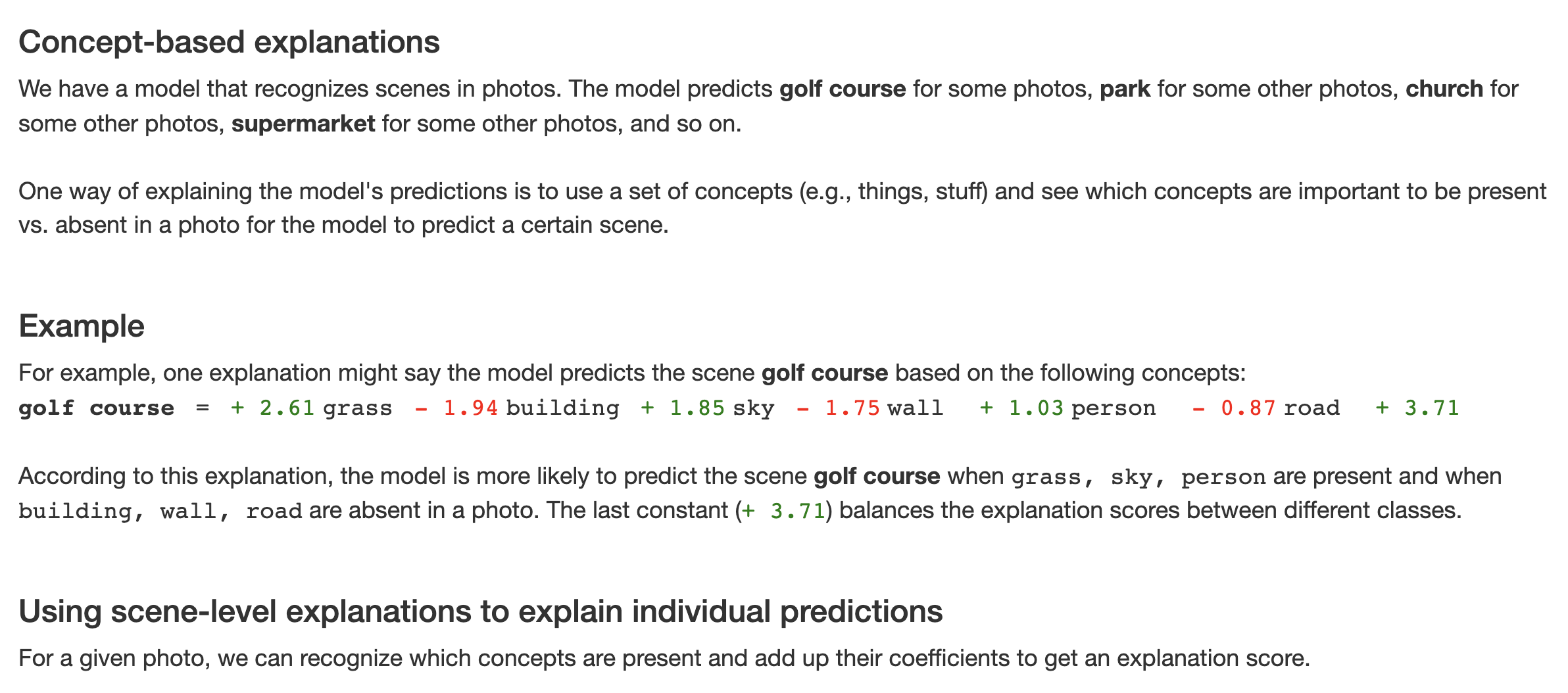}
\includegraphics[width=\linewidth]{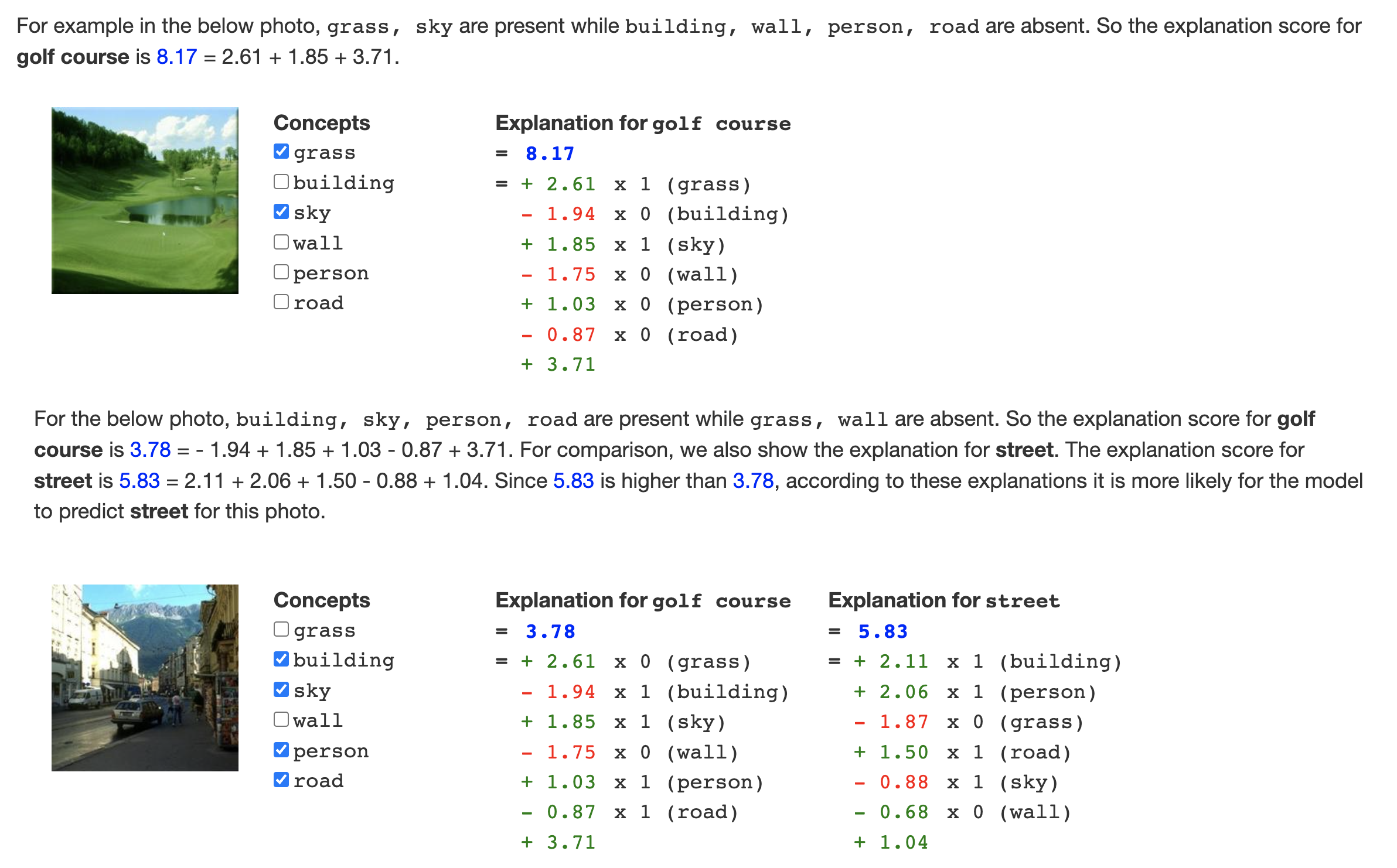}
\caption{\textbf{UI - Method introduction}}
\label{fig:ui_method_intro}
\end{figure*}

\smallsec{Method introduction.} We introduce concept-based explanations in simple terms. This page is not shown for the study with example-based explanations.
See~\cref{fig:ui_method_intro}.

\begin{figure*}[t!]
\centering
\includegraphics[width=\linewidth]{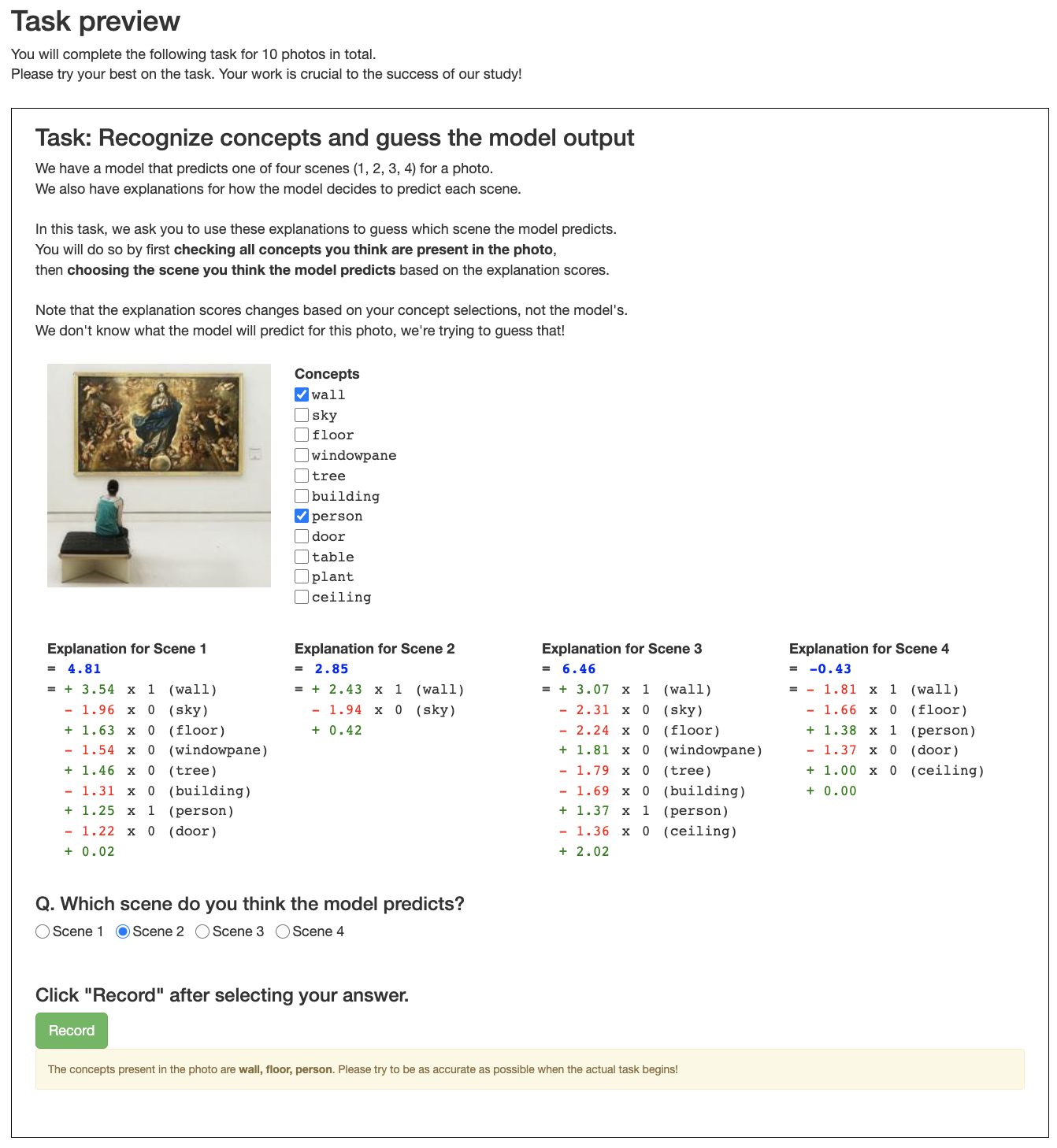}
\caption{\textbf{UI - Task preview}}
\label{fig:ui_task_preview}
\end{figure*}

\smallsec{Task preview } We present a practice example to help participants get familiar with the task. This page is not shown for the study with example-based explanations.
See~\cref{fig:ui_task_preview}.

\begin{figure*}[t!]
\centering
\includegraphics[width=\linewidth]{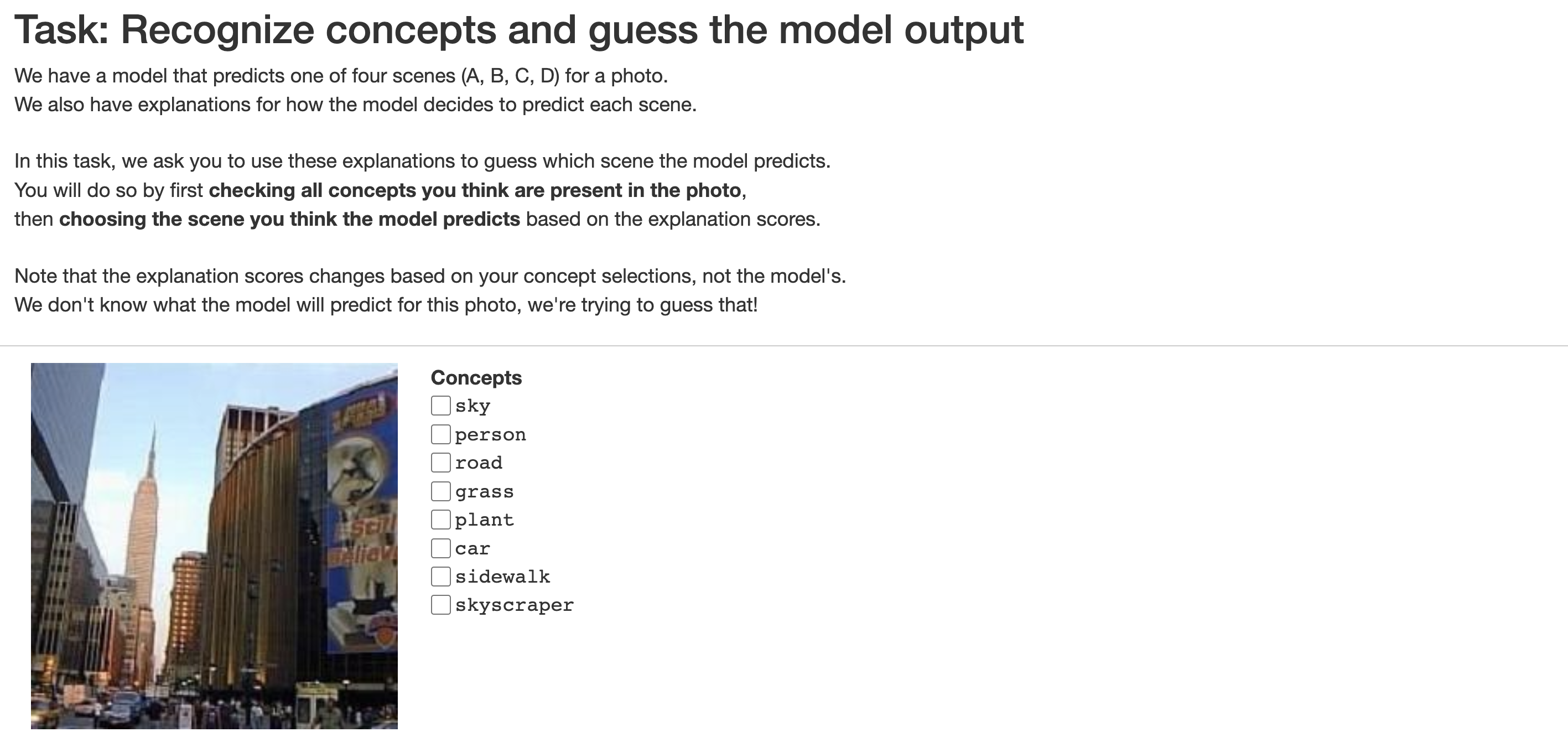}
\includegraphics[width=\linewidth]{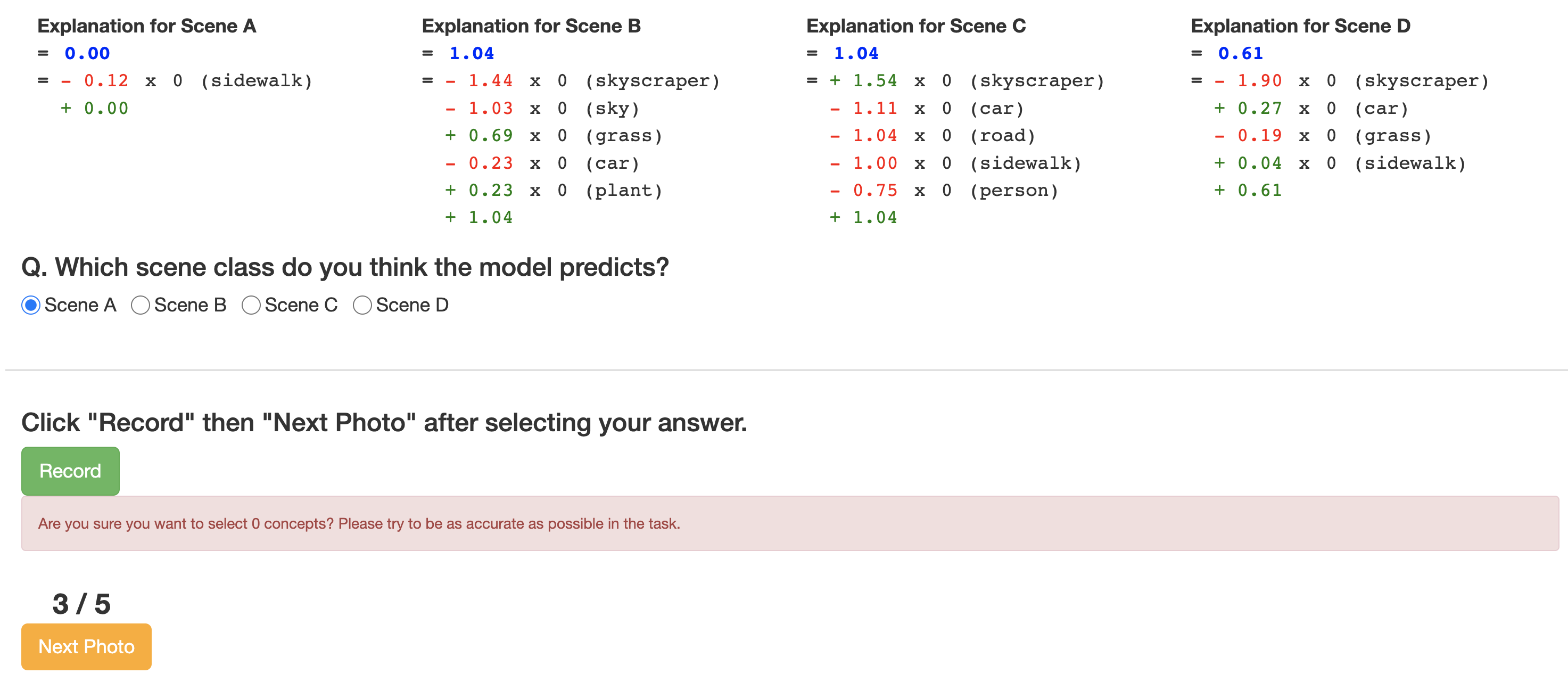}
\caption{\textbf{UI - Part 1: Recognize concepts and guess the model output (8 concepts explanations)}}
\label{fig:ui_task_8concept}
\end{figure*}

\begin{figure*}[t!]
\centering
\includegraphics[width=\linewidth]{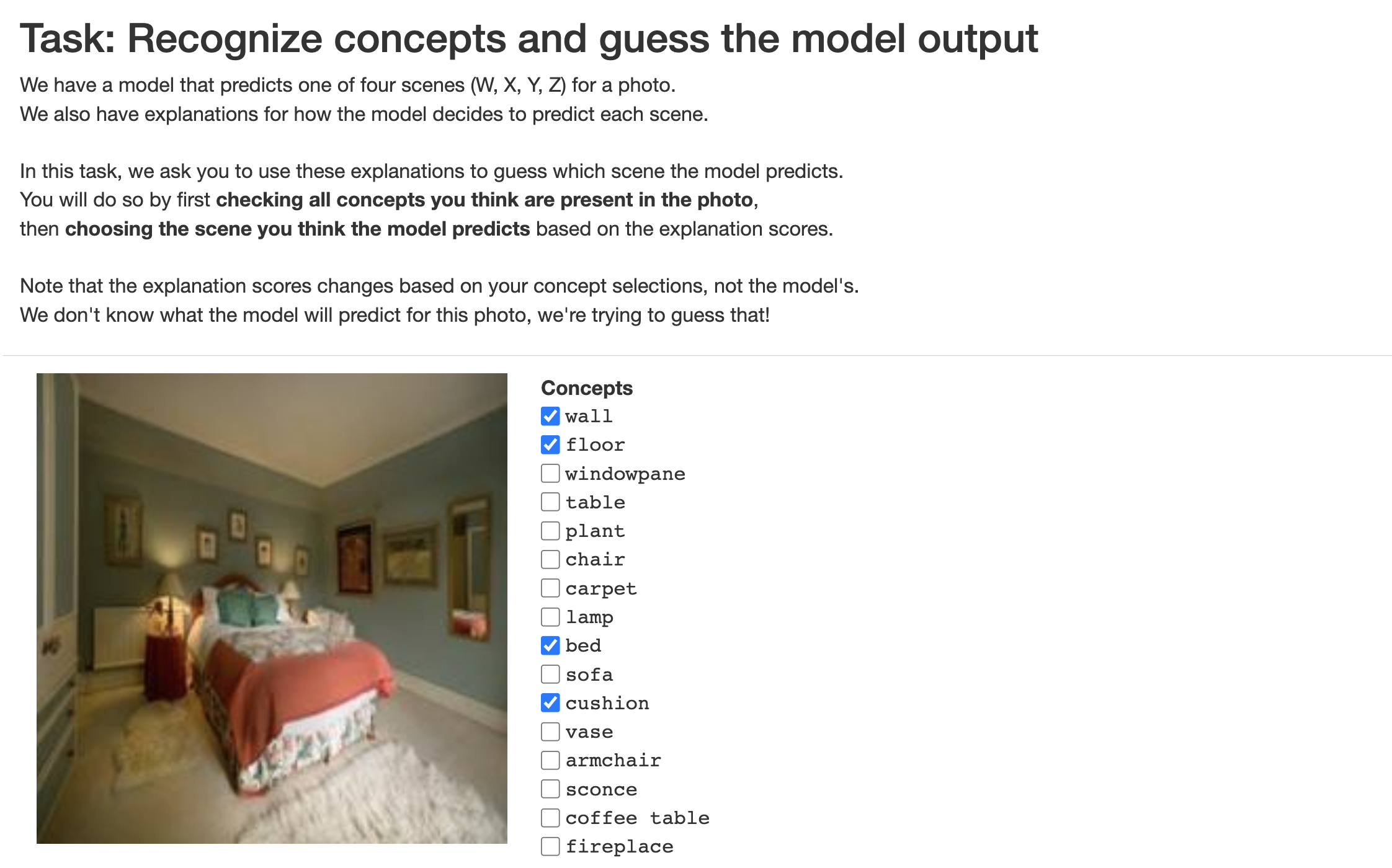}
\includegraphics[width=\linewidth]{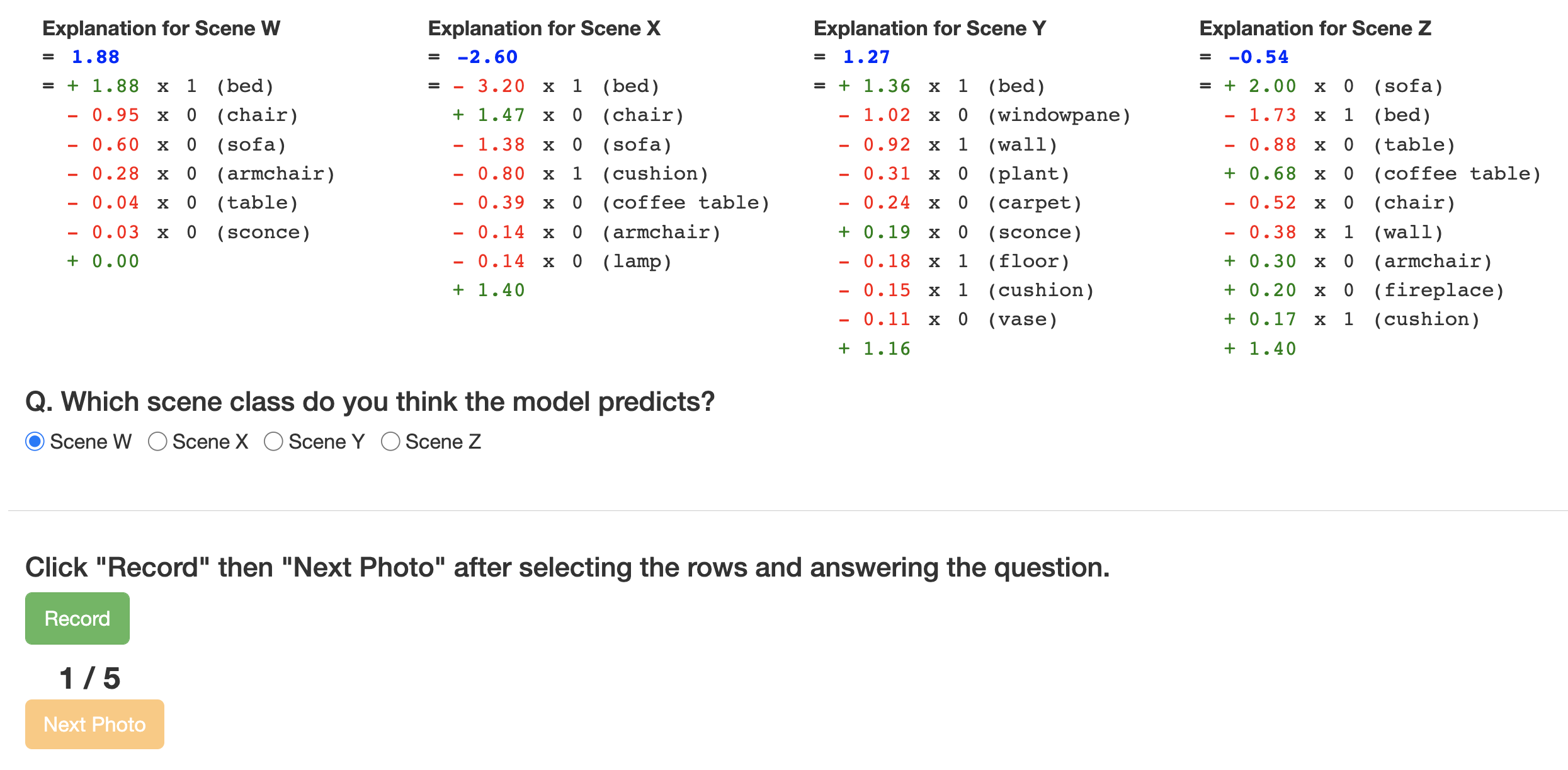}
\caption{\textbf{UI - Part 1: Recognize concepts and guess the model output (16 concepts explanations)}}
\label{fig:ui_task_16concept}
\end{figure*}

\begin{figure*}[t!]
\centering
\includegraphics[width=\linewidth]{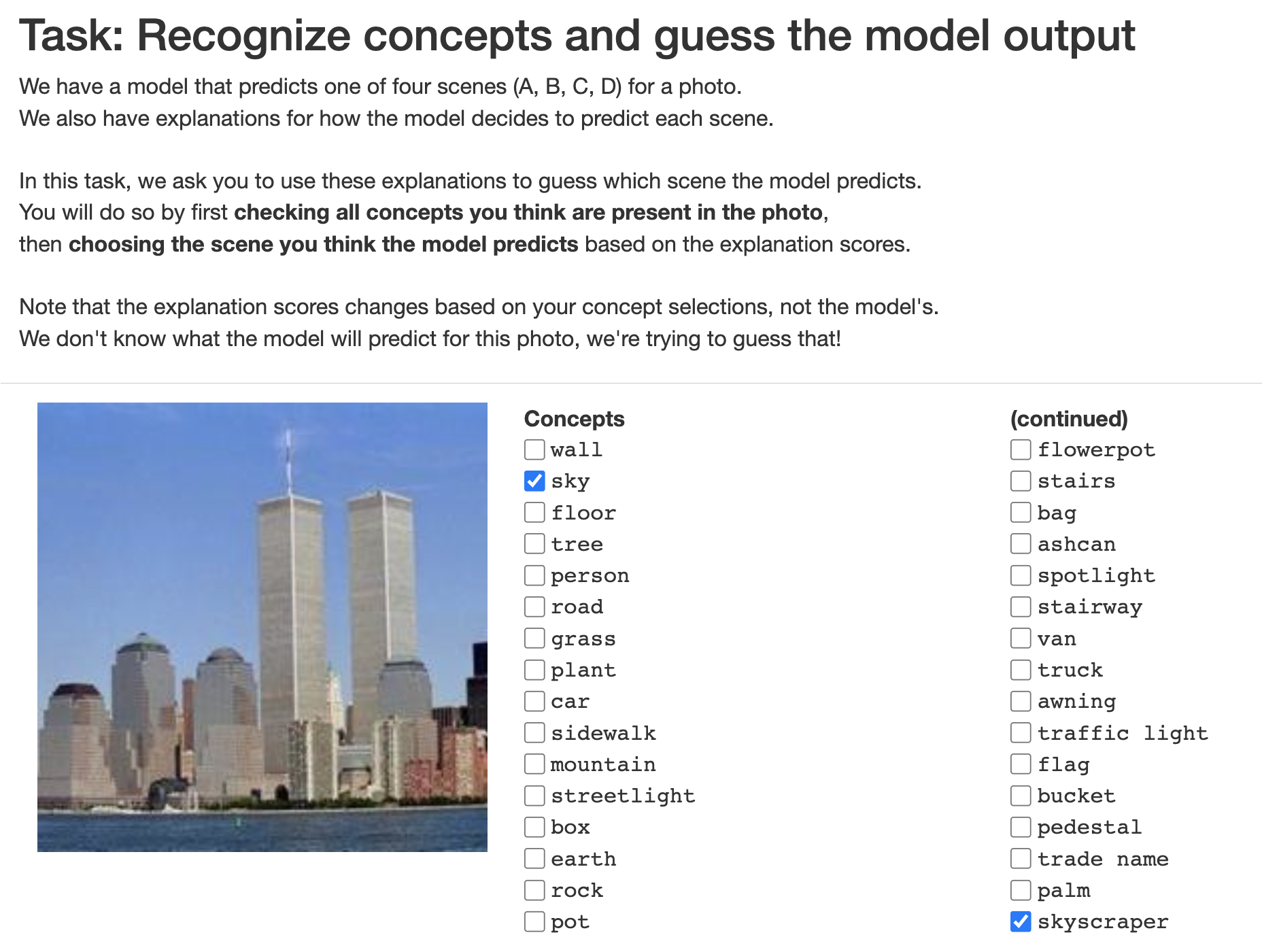}

\vspace{0.5cm}

\includegraphics[width=\linewidth]{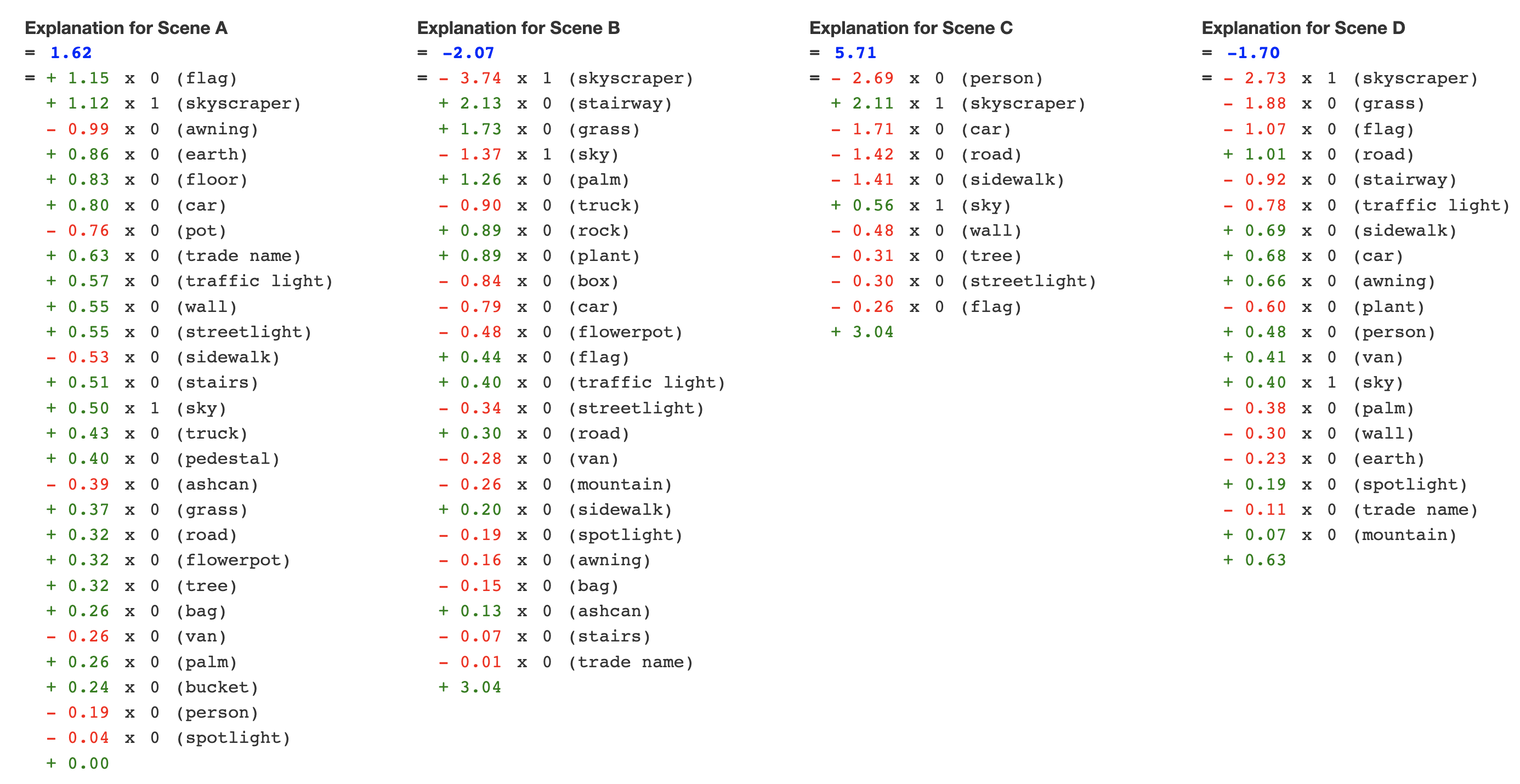}
\includegraphics[width=\linewidth]{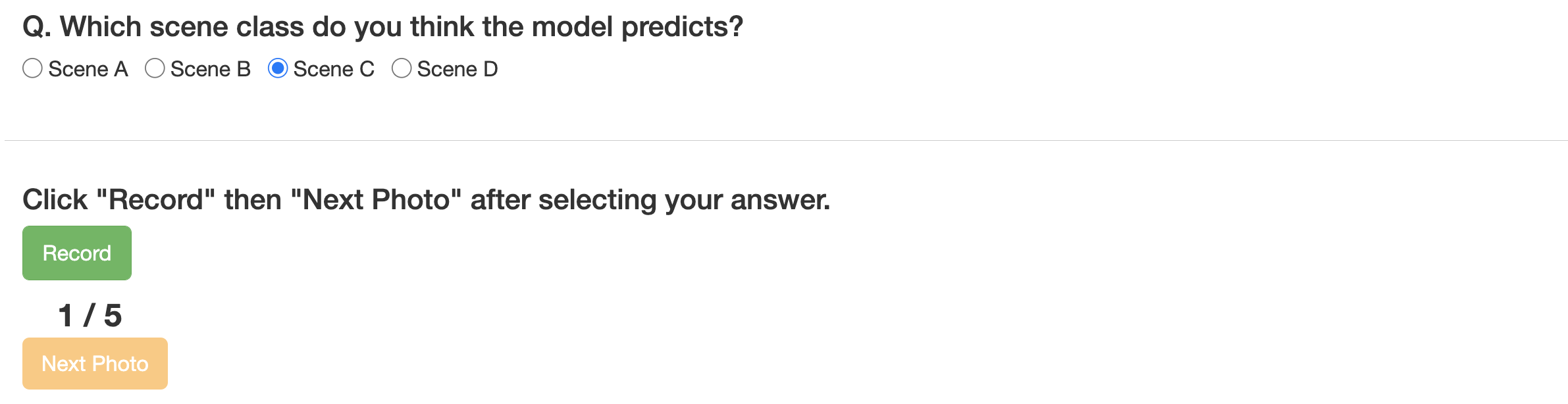}
\caption{\textbf{UI - Part 1: Recognize concepts and guess the model output (32 concepts explanations)}}
\label{fig:ui_task_32concept}
\end{figure*}

\begin{figure*}[t!]
\centering
\includegraphics[width=\linewidth]{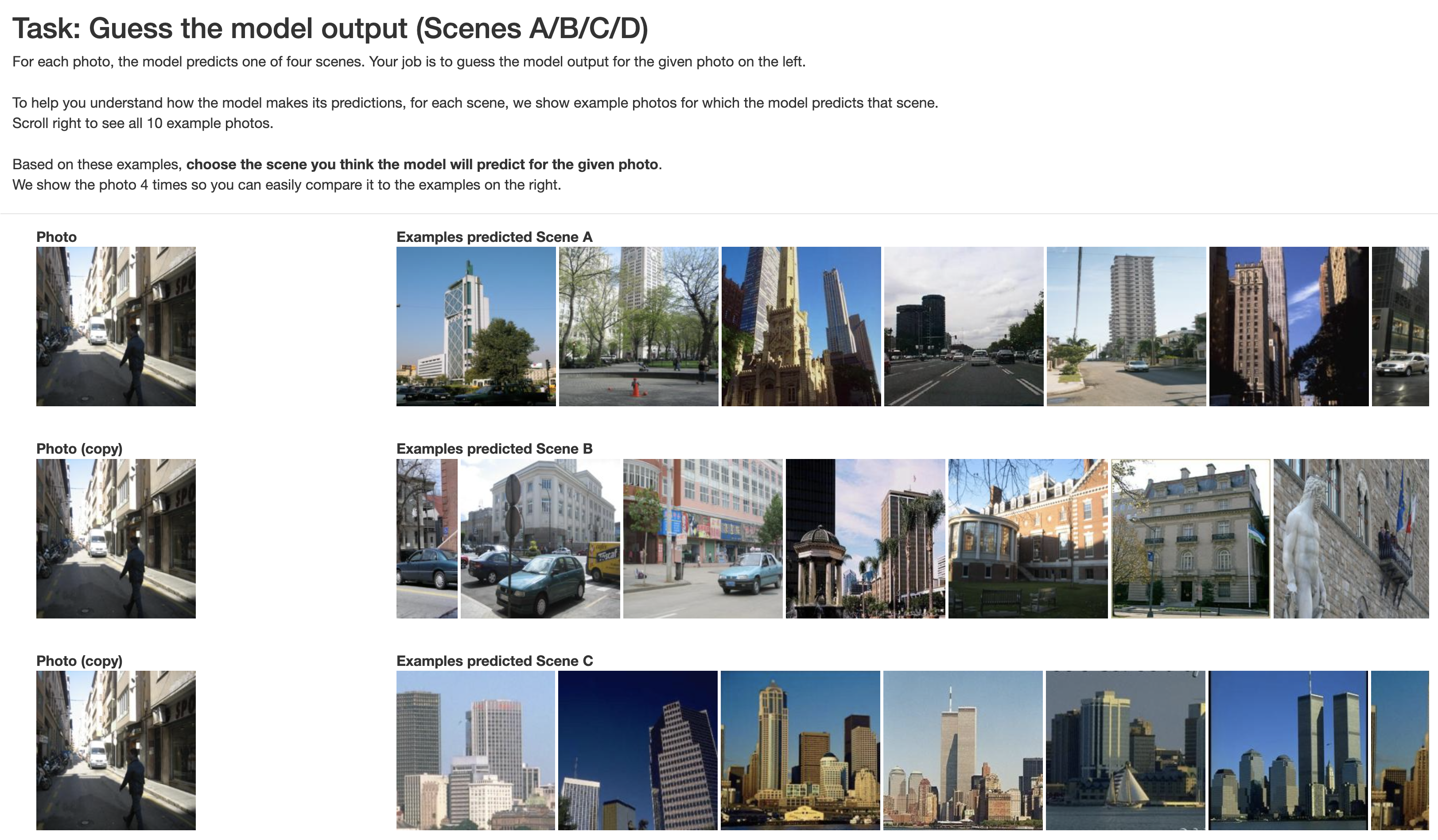}
\includegraphics[width=\linewidth]{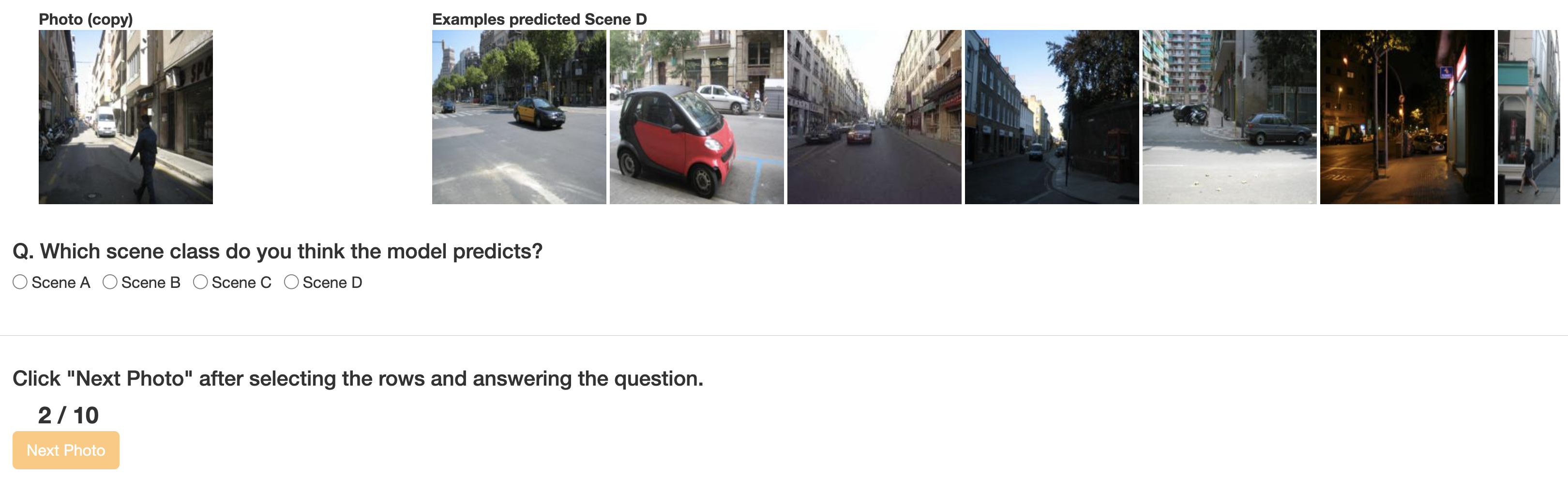}
\caption{\textbf{UI - Part 1: Guess the model output (example-based explanations)}}
\label{fig:ui_task_example}
\end{figure*}

\smallsec{Part 1: Recognize concepts and guess the model output} After the preview, participants move onto the main task where they are asked to recognize concepts in a given photo (for concept-based explanations) and predict the model output (for all explanations). We show the UI for each type of explanation we study:
\begin{itemize}
\item 8 concept explanations (\cref{fig:ui_task_8concept})
\item 16 concepts explanations (\cref{fig:ui_task_16concept})
\item 32 concepts explanations (\cref{fig:ui_task_32concept})
\item Example-based explanations (\cref{fig:ui_task_example})
\end{itemize}

\begin{figure*}[t!]
\centering
\includegraphics[width=\linewidth]{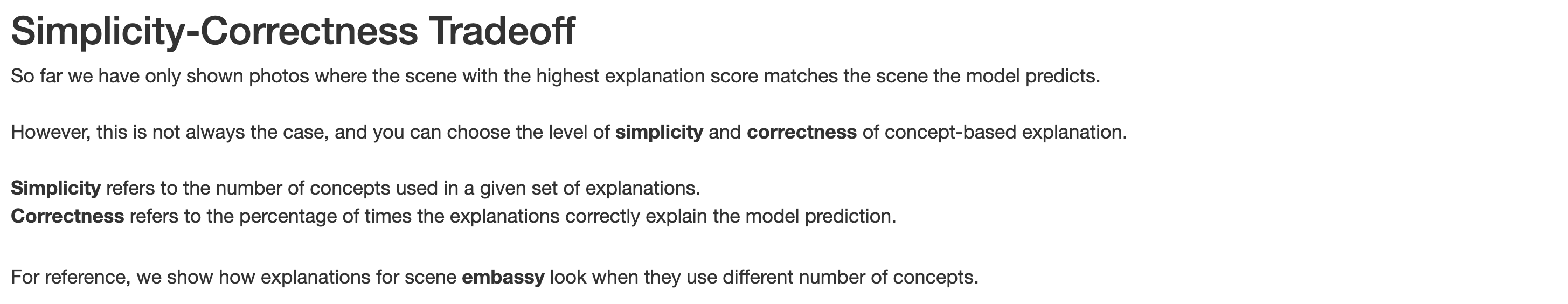}
\includegraphics[width=0.7\linewidth]{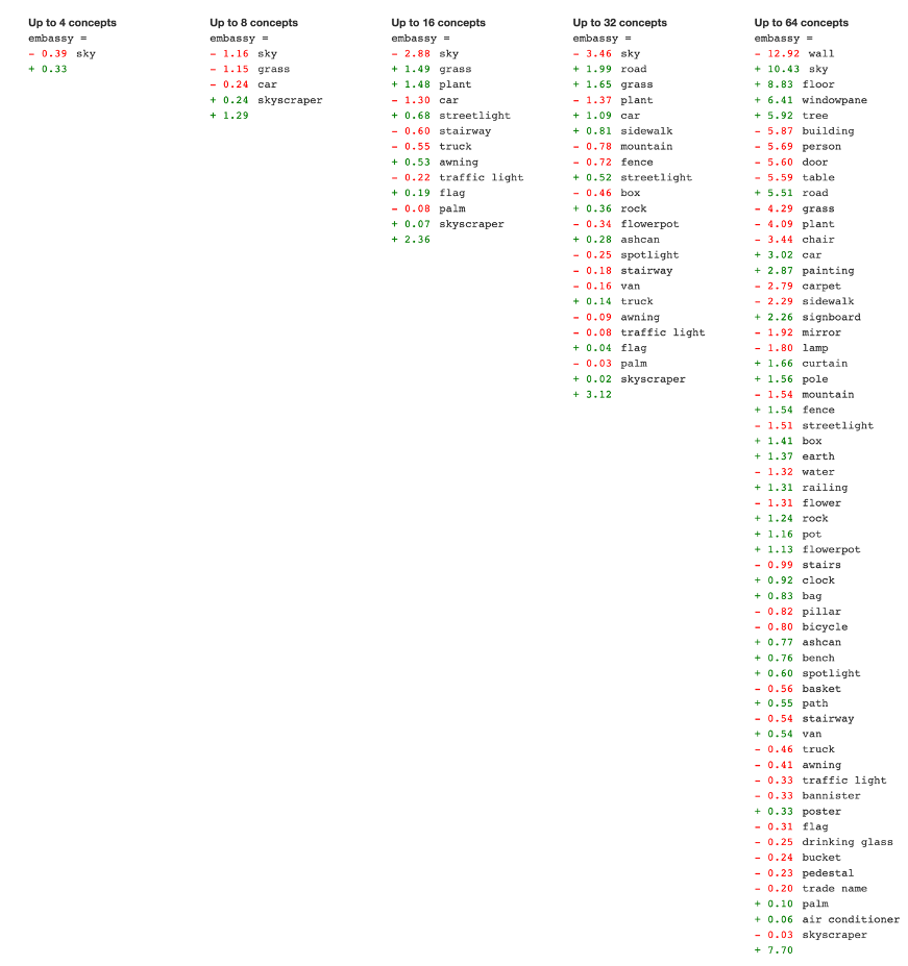}
\includegraphics[width=\linewidth]{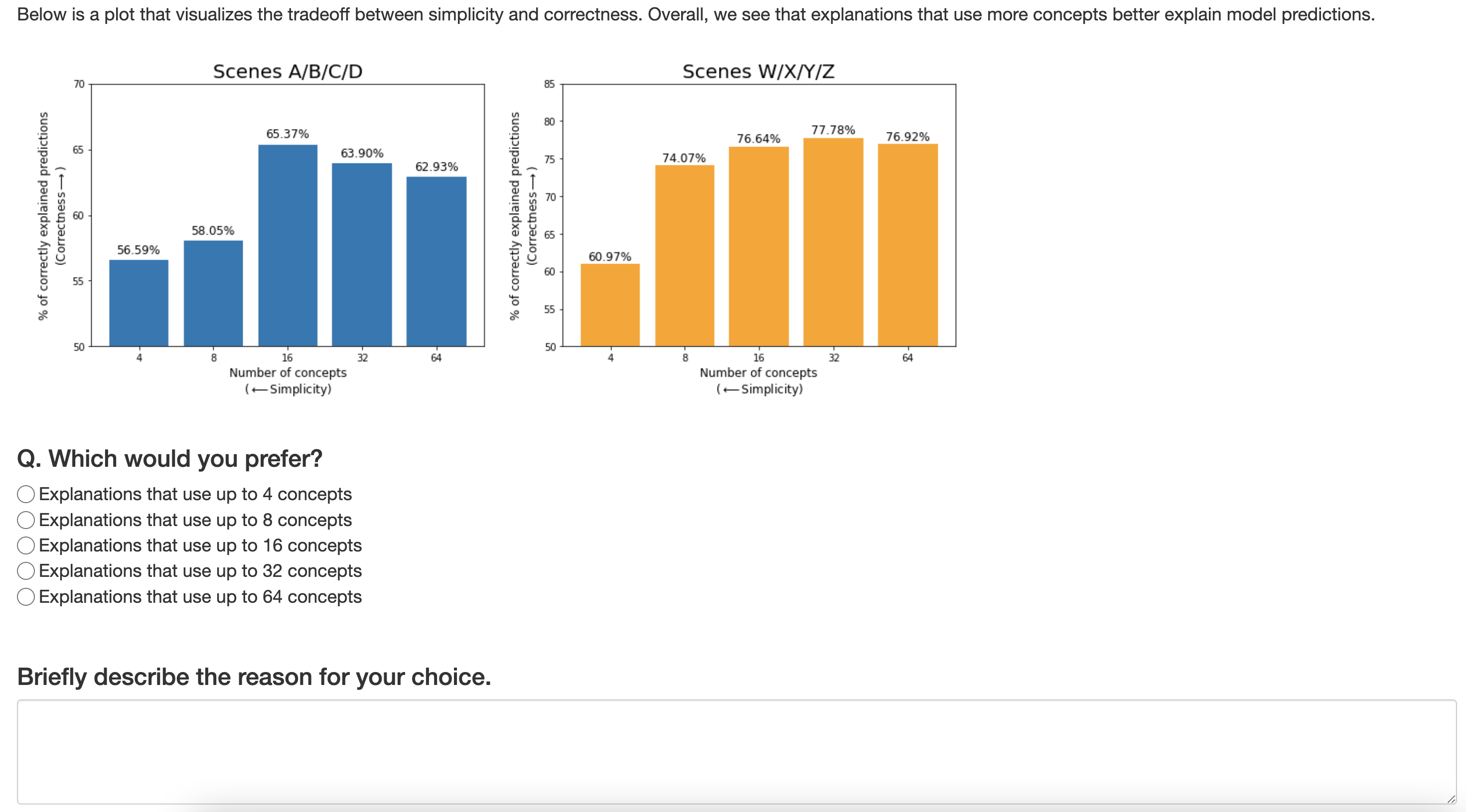}
\caption{\textbf{UI - Part 2: Choose the ideal tradeoff between simplicity and correctness}}
\label{fig:ui_tradeoff}
\end{figure*}

\smallsec{Part 2: Choose the ideal tradeoff between simplicity and correctness.} 
Concept-based explanations can have varying levels of complexity/simplicity and correctness. Hence, we investigate how participants reason with these two properties. To do so, we show examples of concept-based explanations that use different numbers of concepts, as well as bar plots with the correctness values for certain instantiations of concept-based explanations.
We then ask participants to choose the explanation they prefer the most and provide a short written justification for their choice.
See~\cref{fig:ui_tradeoff}.

\begin{figure*}[t!]
\centering
\includegraphics[width=0.85\linewidth]{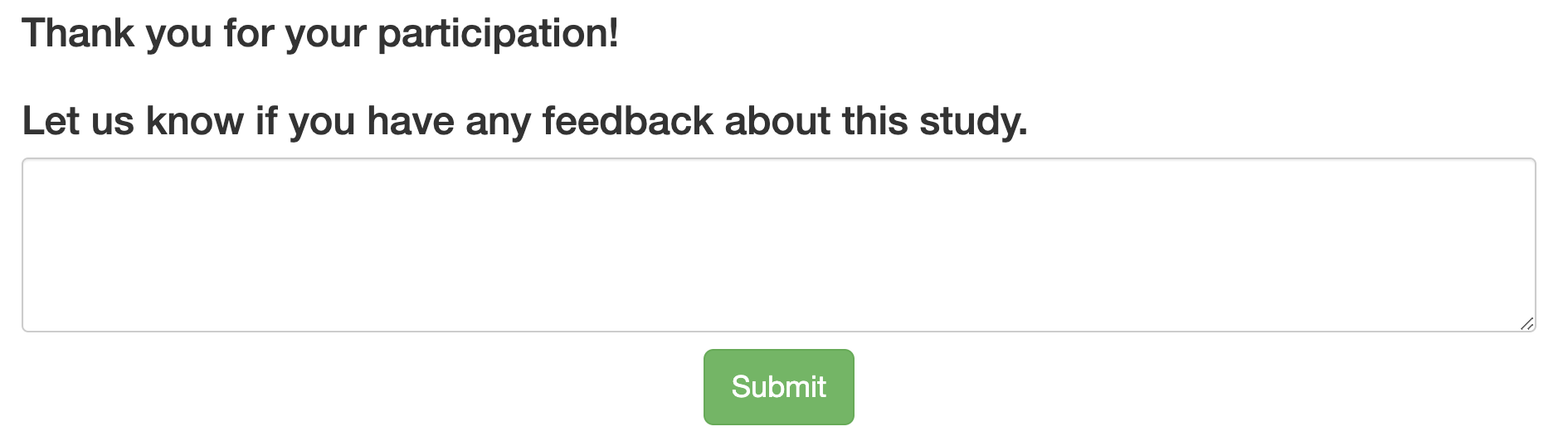}
\caption{\textbf{UI - Feedback}}
\label{fig:ui_feedback}
\end{figure*}

\smallsec{Feedback.} At the end of the study, participants can optionally provide feedback. See~\cref{fig:ui_feedback}.

\clearpage


\end{document}